\documentclass[runningheads]{llncs}

\usepackage{eccv}

\usepackage{eccvabbrv}

\usepackage{graphicx}
\usepackage{booktabs}

\usepackage{wrapfig}

\usepackage[accsupp]{axessibility}  % Improves PDF readability for those with disabilities.

\usepackage{hyperref}

\usepackage{orcidlink}
\usepackage{booktabs}
\usepackage{multirow}

\begin{document}

% ---------------------------------------------------------------
% TODO REVIEW: Replace with your title
\title{JanusMesh: Fast and Zero-Shot 3D Visual Illusion Generation via Cross-Space Denoising} 

% TODO REVIEW: If the paper title is too long for the running head, you can set
% an abbreviated paper title here. If not, comment out.
\titlerunning{JanusMesh}

% TODO FINAL: Replace with your author list. 
% Include the authors' OCRID for the camera-ready version, if at all possible.
\author{Siang-Ling Zhang \and Huai-Hsun Cheng \and Tsung-Ju Yang \and Yu-Lun Liu}

% TODO FINAL: Replace with an abbreviated list of authors.
\authorrunning{S.-L. Zhang et al.}
% First names are abbreviated in the running head.
% If there are more than two authors, 'et al.' is used.

% TODO FINAL: Replace with your institution list.

\institute{National Yang Ming Chiao Tung University, Taiwan}

\maketitle

\begin{figure}[ht]
  \vspace{-6mm}
  \includegraphics[width=\textwidth]{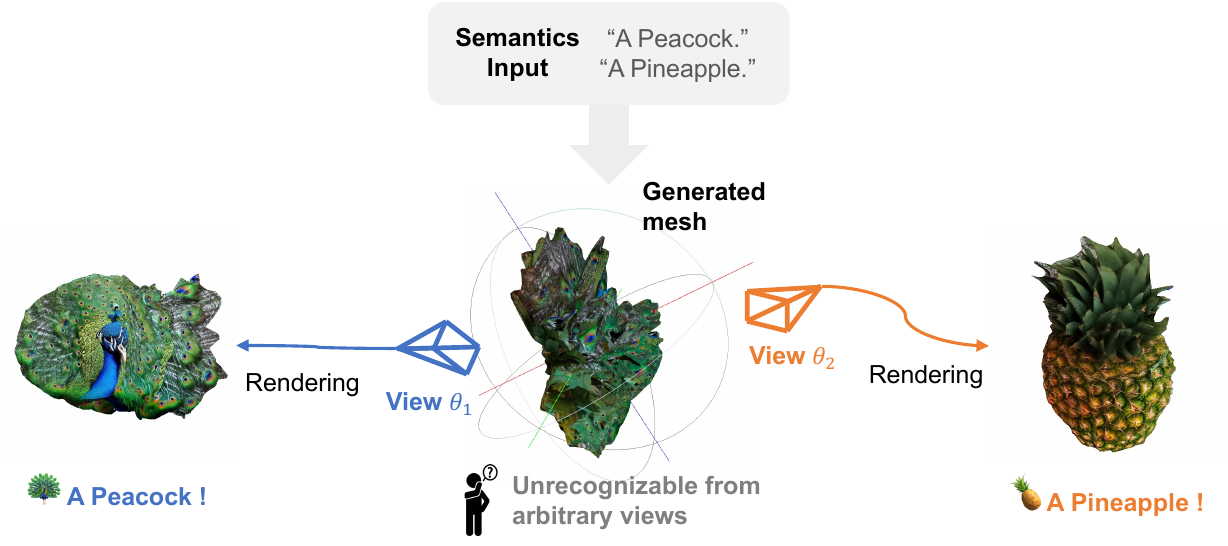}
  \vspace{-6mm}
  \caption{
  \textbf{Zero-shot 3D Visual Illusion Generation.}
  Given two different text prompts, our method creates a single 3D mesh that embodies a dual-semantic visual illusion. The generated shape appears unrecognizable from arbitrary viewpoints, but it perfectly reveals the two target semantics (\emph{e.g.}, a peacock and a pineapple) when observed from specific camera angles (View $\theta_1$ and View $\theta_2$). Our framework achieves this intricate 3D illusion efficiently without requiring per-shape optimization.
  % \textbf{3D Visual Illusion Generation.}
  % Given two text prompts as semantic inputs, our method generates a single 3D mesh that reveals entirely different semantics when viewed from different angles---a peacock from one viewpoint and a pineapple from another.
  % The same mesh appears as a completely different object depending on the observer's viewing angle, achieving a true 3D visual illusion without any additional training or optimization.
  }
  \label{fig:teaser}
  \vspace{-6mm}
\end{figure}

\begin{abstract}
  \vspace{-3mm}
  Creating 3D visual illusions, a single 3D mesh that reveals entirely different semantics from various viewing angles, is a fascinating but tough challenge. Existing optimization-based methods are slow and can produce oversaturated colors. In contrast, naive stitching approaches fail to produce geometrically coherent objects. This results in visible unnatural seams and semantic leaks. In this paper, we present a fast and training-free framework for generating text-driven 3D visual illusions. Our approach decouples the generation into two stages. First, we propose a cross-space dual-branch denoising process. This process dynamically decodes 3D latents into voxel space for CLIP-guided orientation alignment and Signed Distance Field (SDF) blending, which ensures seamless geometric fusion. Second, we introduce a view-conditioned texture synthesis module that projects and aggregates view-specific 2D diffusion priors onto the fused geometry. Extensive experiments demonstrate that our method generates highly realistic, dual-semantic 3D illusions in just 3–5 minutes. It significantly outperforms existing methods in geometric integrity, semantic recognizability, and efficiency. Project page: \url{https://siang1105.github.io/JanusMesh.github.io/}
  \vspace{-3mm}
\end{abstract}

% \section{Introduction}
\vspace{-.5em}
\section{Introduction}
\vspace{-.5em}

\begin{figure}[t]
  \vspace{-1mm}
\includegraphics[width=\columnwidth]{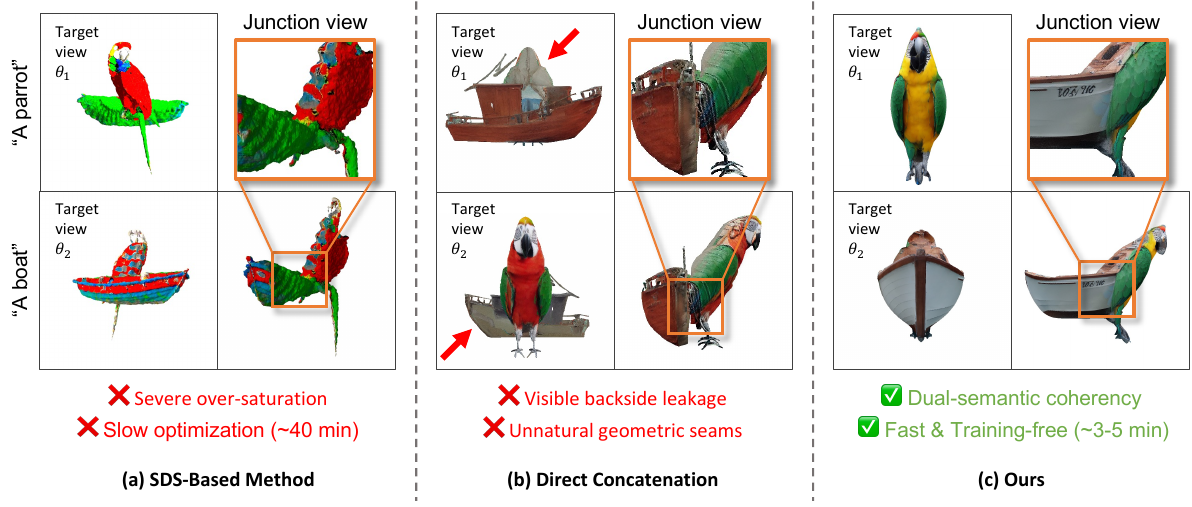}
  \vspace{-6mm}
  \caption{
  \textbf{Comparison of 3D visual illusion generation methods.}
  (a) \textbf{SDS-Based Methods} suffer from severe over-saturation and slow optimization.
  (b) \textbf{Direct Concatenation} exposes unnatural geometric seams and semantic leakage at target views (\textcolor{red}{\textbf{red arrows}}).
  (c) \textbf{Our method} creates a seamless, dual-semantic coherent 3D mesh.
  Unlike previous approaches, our method does not require training.
  It generates high-quality 3D visual illusions in just 3–5 minutes while completely preventing geometric interference between the two semantics.
  % \textbf{Limitations of existing methods for 3D visual illusion generation.}
  % Each column shows results for the prompt pair \textit{``A parrot''} \& \textit{``A boat''}, with target-viewpoint renders on the left and the junction-view render on the right.
  % (a)~\textbf{SDS-based methods} (e.g., Shape From Semantics\cite{li2025shapesemantics3dshape}) optimize each object separately via score distillation sampling, producing severely over-saturated textures; the two objects remain geometrically independent, preventing a coherent 3D illusion.
  % (b)~\textbf{Direct Concatenation} stitches two independently generated objects, exposing a visible geometric seam at the junction where both objects' structures are simultaneously apparent from any viewpoint.
  % (c)~\textbf{Ours} integrates dual semantics into a single coherent geometry via SDF blending in voxel space, presenting clear and recognizable semantics at each target viewpoint while maintaining a natural and complete object appearance.
  }
  \label{fig:motivation}
  \vspace{-3mm}
\end{figure}

Visual illusions have long fascinated human perception by challenging our understanding of physical reality. While diffusion models have enabled computational optical illusions in 2D, bringing this concept into the 3D world---creating a single object that presents entirely different semantics depending on the viewing angle---remains a formidable challenge. As illustrated in Fig.~\ref{fig:teaser}, our goal is zero-shot \emph{3D visual illusion generation}: synthesizing a single 3D mesh that presents, for instance, a peacock from $\theta_1$ and a pineapple from $\theta_2$, while appearing as an abstract geometry from arbitrary angles.

Earlier 3D optical illusion works focused on 2D projections or surface patterns such as shadow art, wireframe silhouettes, or height fields. With text-to-image diffusion models, zero-shot 2D illusion methods (e.g., Visual Anagrams~\cite{geng2024visual}) synchronize noise predictions across views to create multi-interpretation images. Extending this to true 3D meshes, existing approaches rely on SDS to optimize a 3D representation so its renders match different prompts at different viewpoints~\cite{li2025shape}.

However, existing methods fall short in producing high-quality 3D illusions efficiently. As shown in Fig.~\ref{fig:motivation}, optimization-based approaches (e.g., Shape From Semantics~\cite{li2025shape}) require $\sim$40 minutes per shape and suffer from severe color over-saturation. A naive Direct Concatenation of two separately generated objects produces unnatural geometric seams and visible backside leakage, breaking the perceptual illusion.

We present a zero-shot two-stage framework that generates coherent 3D visual illusions in 3--5 minutes. Stage~1 employs a dual-branch denoising process using TRELLIS~\cite{xiang2025structured}, decoding latents into voxel space at each step, aligning objects via CLIP-guided orientation search, and merging them via SDF blending before re-encoding. Stage~2 performs view-conditioned texturing by projecting Stable Diffusion predictions onto the fused mesh. Our method demonstrates superior geometry coherence, texture realism, and semantic recognizability over existing baselines.

In summary, our main contributions are threefold:

\vspace{-2mm}

% \begin{itemize}
%     \item We introduce a zero-shot framework for 3D visual illusion generation, successfully extending generative multi-view illusions from 2D image domains to authentic, fully textured 3D geometric meshes.
%     \item We propose a fast, training-free two-stage architecture. It features a novel cross-space dual-branch denoising strategy with SDF blending and CLIP-guided alignment to ensure geometric integrity, coupled with a view-conditioned texturing process to guarantee dual-semantic coherency.
%     \item We establish a rigorous evaluation protocol (incorporating CLIP, GPT-4o, FID/KID, and novel Object Detection metrics) to assess 3D illusions. Extensive experiments validate that our method largely outperforms existing baselines and naturally scales to generate complex three-object 3D illusions.
% \end{itemize}
\begin{itemize}
    \item A zero-shot framework extending generative multi-view illusions from 2D to fully textured 3D meshes.
    \item A training-free two-stage architecture featuring dual-branch denoising with SDF blending and CLIP-guided alignment for geometric integrity, coupled with view-conditioned texturing for dual-semantic coherency.
    \item A rigorous evaluation protocol incorporating CLIP, GPT-4.1-mini, FID/KID, and a novel Object Detection metric, with experiments validating our method over baselines and demonstrating scalability to three-object illusions.
\end{itemize}
\vspace{-4mm}
\vspace{-.5em}
\section{Related Work}
\vspace{-.5em}
\subsubsection{Computational Optical Illusions.}
Prior work has explored appearance-varying 3D objects, including shadow art~\cite{mitra2009shadow,sadekar2022shadow,wang2024neural}, wire art~\cite{hsiao2018multi,qu2024wired}, view-dependent heightfields~\cite{perroni2023constructing}, SDF-based anamorphic packing~\cite{debnath2025rasp}, and spatially ambiguous geometry~\cite{dodik2025meschers}; in 2D, spatial-frequency decomposition~\cite{oliva2006hybrid,geng2024factorized} shifts perceived content with viewing distance, while progressive vector sketching~\cite{cheng2026stroke} transforms perceived semantics through sequential stroke addition. These approaches produce 2D projections, surface patterns, or---as with adversarially injected viewpoint-dependent content in 3D Gaussian Splatting~\cite{ke2025stealthattack}---view-specific artifacts, whereas our work generates a fully textured 3D mesh with distinct semantics from different viewpoints.
\vspace{-.5em}
\subsubsection{Illusion Generation with Diffusion Models.}
Text-to-image diffusion models~\cite{rombach2022high,ho2020denoising,song2020denoising} have enabled new illusion synthesis. SDS-based methods~\cite{burgert2024diffusion,poole2022dreamfusion} optimize for multiple prompts but converge slowly. Visual Anagrams~\cite{geng2024visual} introduced zero-shot illusions via per-view noise averaging, extended to frequency decompositions~\cite{geng2024factorized}, phase-transfer~\cite{gao2025ptdiffusion}, and audio-visual spectrograms~\cite{chen2024images}; in 3D, Illusion3D~\cite{feng2024illusion3d} and LookingGlass~\cite{chang2025lookingglass} lift these priors into NeRF and anamorphic images, respectively. Our work extends the zero-shot spirit of~\cite{geng2024visual} to native 3D latent space, producing a fully textured mesh.
\vspace{-.5em}
\subsubsection{Text-to-3D Generation.}
Optimization-based methods~\cite{poole2022dreamfusion,lin2023magic3d,wang2023prolificdreamer,chen2023fantasia3d} distill 2D diffusion priors via SDS, with improvements via interval score matching~\cite{liang2024luciddreamer}, rectified-flow distillation~\cite{yang2024text}, and Gaussian acceleration~\cite{tang2023dreamgaussian,yi2024gaussiandreamer}. Feed-forward methods combine multi-view diffusion~\cite{liu2023zero,liu2023syncdreamer,long2024wonder3d,shi2023mvdream} with reconstruction networks~\cite{hong2023lrm,xu2024instantmesh,wang2024crm}. Application-driven variants extend generation to scene scale and alternative styles: training-free pipelines complete fully textured room-scale meshes from sparse images~\cite{li2024genrc}, diffusion priors enable city-scale 3D scene creation with iterative geometry and texture refinement~\cite{lee2025skyfall}, and differentiable mesh optimization under 2D pixel-art supervision produces stylized 3D content~\cite{huang2026voxify3d}. Native 3D generative models~\cite{jun2023shap,zhao2023michelangelo,wu2024direct3d,lan2025gaussiananything,li2025triposg} learn latent representations directly from 3D data; TRELLIS~\cite{xiang2025structured} encodes geometry and appearance in a sparse structured latent space via rectified flow~\cite{liu2022flow}, which we repurpose for dual-semantic mesh generation.
\vspace{-.5em}
\subsubsection{Synchronized Diffusion Denoising.}
Merging denoising trajectories enables compositional generation~\cite{liu2022compositional}, seamless panoramas~\cite{bar2023multidiffusion}, and perceptual synchronization~\cite{lee2023syncdiffusion}, with advanced samplers addressing compositional failures~\cite{du2023reduce,kim2023collaborative}. Recent works average clean-image predictions rather than noise~\cite{kim2024synctweedies} and apply spatial guidance in 3D latents~\cite{fedele2025spacecontrol}. Closely related, multi-view diffusion outpainting coordinates denoising across views with geometry-aware strategies for sparse-view reconstruction~\cite{huang2025gamo}, and 3D-aware $360^\circ$ video diffusion decouples texture refinement from a 3D cache that enforces geometric consistency~\cite{chen2026pantheon360}. Unlike these consistency-enforcing frameworks, our dual-branch denoising uniquely enforces divergent semantics at target viewpoints via SDF fusion.
\vspace{-.5em}
\subsubsection{3D Texture Synthesis.}
Diffusion-based mesh texturing spans depth-conditioned inpainting~\cite{richardson2023texture}, ControlNet-enhanced generation~\cite{chen2023text2tex,zeng2024paint3d}, multi-view consistent UV-latents~\cite{cao2023texfusion,liu2024text,cheng2025mvpaint}, and single-pass feed-forward models~\cite{yu2024texgen} with material~\cite{deng2024flashtex} or appearance~\cite{yeh2024texturedreamer} extensions. Crucially, all these methods apply a single prompt uniformly. Our view-conditioned synthesis instead assigns distinct prompts to different angular sectors, back-projecting viewpoint-specific clean images via cosine-weighted blending.
\vspace{-.5em}
\subsubsection{CLIP-Guided 3D Understanding and Generation.}
CLIP's~\cite{radford2021learning} render-caption similarity enables zero-shot 3D generation~\cite{jain2022zero}, mesh stylization~\cite{michel2022text2mesh}, NeRF manipulation~\cite{wang2022clip, jain2021putting}, and latent score distillation~\cite{metzer2023latent}. Leveraging this render-and-score paradigm, our CLIP-guided Orientation Search automatically selects the relative rotation that maximizes silhouette alignment between the two objects. This resolves geometric mismatches that would otherwise cause SDF fusion failures.

\vspace{-.5em}
\section{Method}

\vspace{-.5em}
\subsection{Preliminaries}

\vspace{-.5em}
\subsubsection{TRELLIS: Structured 3D Latent Representation.}
TRELLIS~\cite{xiang2025structured} is a two-stage Rectified Flow~\cite{lipman2022flow} 3D generator that first predicts a low-resolution sparse voxel structure, then refines it with high-dimensional appearance features. We perform geometry blending specifically during this first structural stage. Because direct spatial transformations distort latent distributions, we draw inspiration from LookingGlass~\cite{chang2025lookingglass}: at each denoising step, we decode the latent into voxel space, perform SDF blending, and re-encode the fused result back to the latent space. This cross-space denoising ensures the geometric validity of the blended mesh.

\vspace{-.5em}
\subsubsection{Visual Anagrams.}
Visual Anagrams~\cite{geng2024visual} generates 2D illusions by averaging noisy estimates across transformed views in a canonical space. Because this restricts transformations to be orthogonal, SyncTweedies~\cite{kim2024synctweedies} relaxes the constraint by averaging estimated clean latents instead: $\hat{\mathbf{z}}_{1|t}^i = \pi_i ( \frac{1}{N} \sum_j \pi_j^{-1} ( \mathbf{z}_{1|t}^j ) )$. We extend this synchronization principle from 2D pixels to 3D voxels, blending predicted clean geometries via SDF averaging to generate view-dependent dual-semantic meshes.
\vspace{-.5em}
\subsection{Overview}
\vspace{-.5em}
%%%%%%%%%%%%%%%%%%%%%%%%%%%%%%%%%%%%%%%%%%%%%%%%%%

\begin{figure*}[t]
\vspace{-1mm}
  \includegraphics[width=\textwidth]{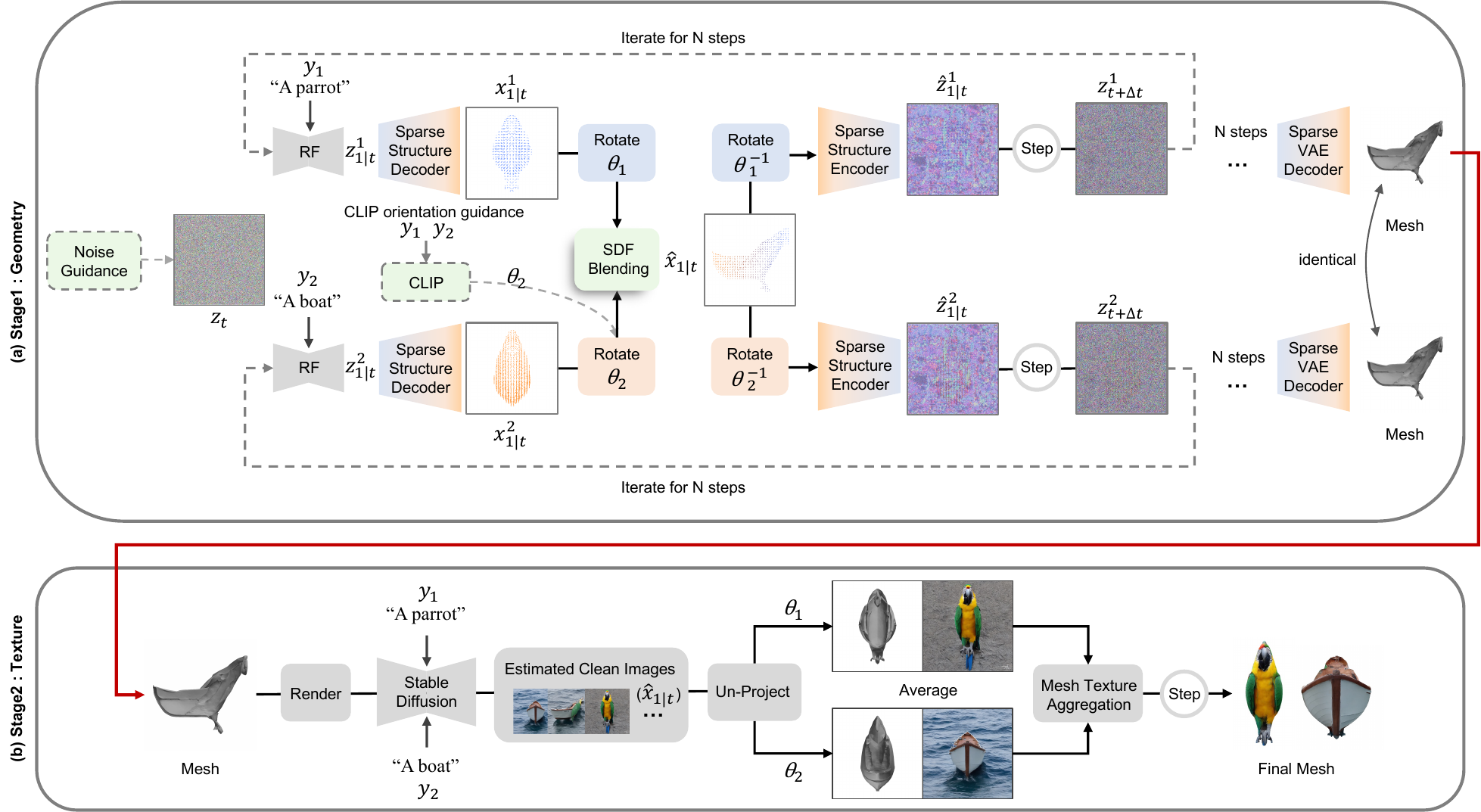}
  \vspace{-6mm}
  \caption{
  \textbf{Pipeline overview.}
  \textbf{(a) Stage~1} employs dual-branch denoising. At each step, latents are decoded to voxel space, rotation-aligned, and fused via SDF blending (Fig.~\ref{fig:sdf_blend}), then re-encoded to continue denoising, producing a single unified 3D mesh.
  \textbf{(b) Stage~2} applies view-conditioned texturing to the fused mesh. Estimated clean images $\hat{x}_{1|t}$ are predicted via Stable Diffusion, un-projected from viewpoints $\theta_1$ and $\theta_2$, and iteratively aggregated via Mesh Texture Aggregation, producing a single object with distinct semantics at each target viewpoint.
  }
  \vspace{-2mm}
  \label{fig:pipeline}
  \vspace{-3mm}
\end{figure*}

% We propose a two-stage framework for generating 3D visual illusion objects that present different semantics when viewed from different angles. Given two text prompts $y_1$ and $y_2$ and their corresponding target viewpoints $\theta_1$ and $\theta_2$, our goal is to generate a single 3D mesh that presents the semantics of $y_1$ when observed from $\theta_1$, and the semantics of $y_2$ when observed from $\theta_2$.

% A successful 3D visual illusion must simultaneously satisfy three criteria: (1)~\textbf{Semantic Recognizability} --- the object clearly presents the intended semantics at each target viewpoint; (2)~\textbf{Geometric Integrity} --- the object appears as a single coherent geometry from any viewpoint, rather than two separate objects; and (3)~\textbf{Illusion Effect} --- at non-target viewpoints, the object does not obviously reveal geometric cues of the opposing semantic, producing a natural visual reversal as the observer rotates their perspective.

% As illustrated in Fig.~\ref{fig:pipeline}, Stage~1 generates the blended geometry via dual-branch denoising in latent space(Sec.~\ref{sec:stage1}), while Stage~2 applies view-conditioned textures to the fused mesh according to each target viewpoint(Sec.~\ref{sec:stage2}).

Given prompts $y_1, y_2$ and target viewpoints $\theta_1, \theta_2$, our framework generates a single 3D mesh exhibiting $y_1$ at $\theta_1$ and $y_2$ at $\theta_2$. A successful illusion requires: (1)~\textbf{Semantic Recognizability} at target views; (2)~\textbf{Geometric Integrity} from any viewpoint; and (3)~\textbf{Illusion Effect} (concealing opposing semantics at non-target views), all evaluated in Sec.~\ref{sec:metrics}. As shown in Fig.~\ref{fig:pipeline}, our pipeline achieves this via Stage~1 dual-branch geometric denoising (Sec.~\ref{sec:stage1}) and Stage~2 view-conditioned texturing (Sec.~\ref{sec:stage2}).
% Given two text prompts $y_1$ and $y_2$ with target viewpoints $\theta_1$ and $\theta_2$, our two-stage framework generates a single 3D mesh presenting the semantics of $y_1$ at $\theta_1$ and $y_2$ at $\theta_2$. A successful illusion must simultaneously satisfy three criteria: (1)~\textbf{Semantic Recognizability} --- clear intended semantics at each target viewpoint; (2)~\textbf{Geometric Integrity} --- a single coherent geometry from any viewpoint; and (3)~\textbf{Illusion Effect} --- no obvious geometric cues of the opposing semantic at non-target viewpoints. These criteria are quantitatively evaluated in Sec.~\ref{sec:metrics}. As illustrated in Fig.~\ref{fig:pipeline}, Stage~1 generates blended geometry via dual-branch denoising (Sec.~\ref{sec:stage1}), while Stage~2 applies view-conditioned textures to the fused mesh (Sec.~\ref{sec:stage2}).

%%%%%%%%%%%%%%%%%%%%%%%%%%%%%%%%%%%%%%%%%%%%%%%%%%
\vspace{-.5em}
\subsection{Dual-Branch Geometry Generation (Stage~1: Geometry)}
%%%%%%%%%%%%%%%%%%%%%%%%%%%%%%%%%%%%%%%%%%%%%%%%%%
\label{sec:stage1}

\vspace{-.5em}
\subsubsection{Dual-Branch Denoising and Clean Latent Estimation.}

% Stage~1 builds on TRELLIS's Rectified Flow, starting from a shared initial noise $z_t$ and performing two independent denoising branches conditioned on $y_1$ and $y_2$ respectively, estimating the clean latent at each timestep $t$:

Stage~1 builds on TRELLIS's Rectified Flow, starting from a shared initial noise $z_t$ and performing two independent denoising branches conditioned on $y_1$ and $y_2$, estimating the clean latent at each timestep $t$:
\begin{equation} \small
    x^1_{1|t} = z_t + u_\theta(z_t;\, t,\, y_1)(1 - t),
\end{equation}
\begin{equation}\small
    x^2_{1|t} = z_t + u_\theta(z_t;\, t,\, y_2)(1 - t),
\end{equation}
where $u_\theta$ is the Rectified Flow Network. We apply Classifier-Free Guidance (CFG)~\cite{ho2022classifier} with Interval CFG to improve generation quality while avoiding over-saturation at extreme noise levels.

% where $u_\theta$ is the Rectified Flow Network. We apply Classifier-Free Guidance (CFG)~\cite{ho2022classifier} with guidance scale $\omega = 7.5$, activated only within the interval $t \in [0.5, 0.95]$ (Interval CFG), to improve generation quality while avoiding over-saturation at extreme noise levels.

\vspace{-.5em}
\subsubsection{Geometry Blending in Voxel Space.}

\begin{figure}[t]
\vspace{-2mm}
  \centering
  \includegraphics[width=0.8\columnwidth]{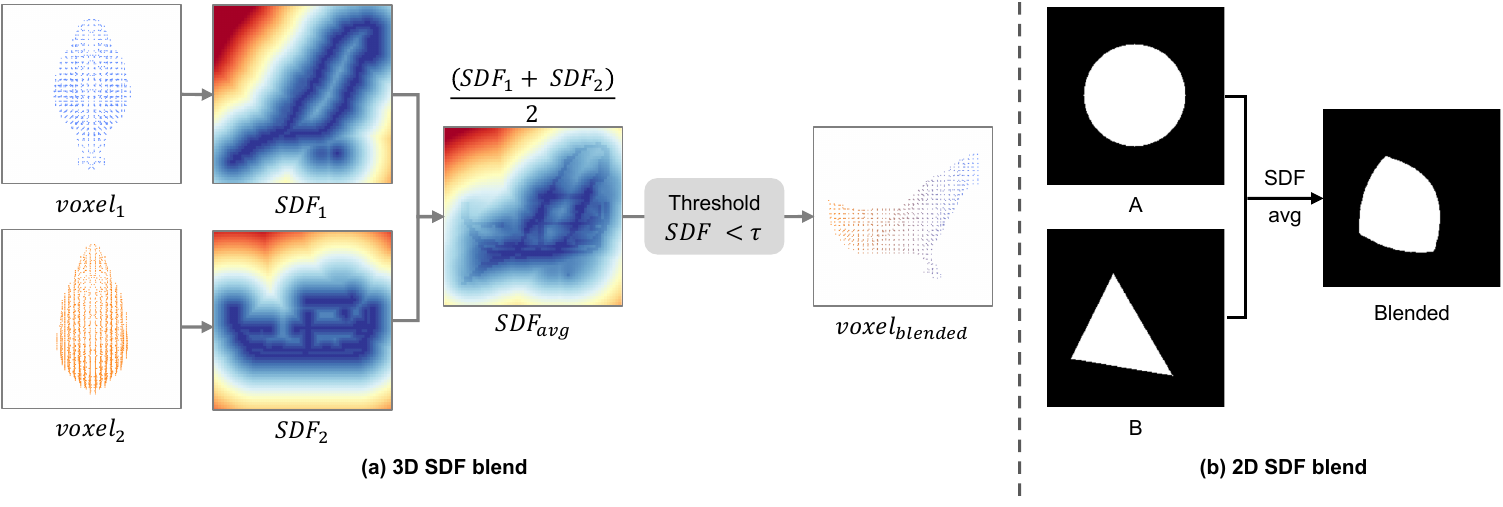}
  % \caption{
  % \textbf{SDF blending.}
  % (a)~Given two rotation-aligned voxels $\text{voxel}_1$ and $\text{voxel}_2$, we compute their signed distance fields $\text{SDF}_1$ and $\text{SDF}_2$, take their element-wise average $\text{SDF}_\text{avg} = (\text{SDF}_1 + \text{SDF}_2) / 2$, and binarize the result with threshold $\tau$ to obtain $\text{voxel}_\text{blended}$.
  % (b)~A 2D illustration of the blending process: averaging the SDFs of a circle~(A) and a triangle~(B) naturally produces a blended shape that lies geometrically between the two, forming a smooth intermediate contour.
  % }
  \vspace{-3mm}
  \caption{
  \textbf{SDF blending.}
  (a)~Given two rotation-aligned voxels, we compute their SDFs, take the element-wise average, and binarize with threshold $\tau$ to obtain the blended voxel.
  (b)~2D illustration: averaging the SDFs of a circle~(A) and a triangle~(B) produces a smooth intermediate contour lying geometrically between the two.
  }
  \vspace{-3mm}
  \label{fig:sdf_blend}
  \vspace{-3mm}
\end{figure}

% The clean latent estimates are decoded into occupancy grids via the Sparse VAE Decoder, yielding the voxel representations $V_1$ and $V_2$ of the two objects. To ensure that the final object presents semantic $y_1$ at angle $\theta_1$ and semantic $y_2$ at angle $\theta_2$, we rotate $V_2$ by $\theta_2$ to align both voxels in a common reference frame.

% Since directly averaging binary occupancy grids lacks geometric continuity, we convert both voxels into Signed Distance Fields (SDFs), take their element-wise average, and binarize the result with a fixed threshold $\tau = 0.8$ to obtain the blended geometry estimate $\hat{x}_{1|t}$:

The clean latent estimates $x^1_{1|t}$ and $x^2_{1|t}$ are decoded via the Sparse Structure Decoder into voxel representations $v_1$ and $v_2$. We rotate $v_2$ by $\theta_2$ to align both voxels in a common reference frame. Since directly averaging binary occupancy grids lacks geometric continuity, we convert both into Signed Distance Fields (SDFs), average them element-wise, and binarize with threshold $\tau$ to obtain the blended geometry $\hat{x}_{1|t}$:
\begin{equation}\small
    \text{SDF}_\text{blend} = \frac{\text{SDF}(v_1) + \text{SDF}(v_2)}{2},
\end{equation}
\begin{equation}\small
    \hat{x}_{1|t} = [\text{SDF}_\text{blend} < \tau].
\end{equation}
The blending process is illustrated in Fig.~\ref{fig:sdf_blend}. After blending, $\hat{x}_{1|t}$ is rotated by $-\theta_2$ to restore $v_2$'s original coordinate frame, then re-encoded into $\hat{z}^1_{1|t}$ and $\hat{z}^2_{1|t}$ via the Sparse Structure Encoder for the next denoising step.

% The SDF blending process is illustrated in Fig.~\ref{fig:sdf_blend}. After blending, $\hat{x}_{1|t}$ is rotated by $-\theta_2$ to restore it to $V_2$'s original coordinate frame. The blended geometry is then re-encoded into latent representations $\hat{z}^1_{1|t}$ and $\hat{z}^2_{1|t}$ via the Sparse VAE Encoder for the next denoising step. The entire denoising and blending loop is executed for 25 steps, with one geometry blending operation per step, ultimately producing the fused 3D mesh.

%%%%%%%%%%%%%%%%%%%%%%%%%%%%%%%%%%%%%%%%%%%%%%%%%%
\vspace{-.5em}
\subsection{View-Conditioned Texture Synthesis (Stage~2: Texture)}
%%%%%%%%%%%%%%%%%%%%%%%%%%%%%%%%%%%%%%%%%%%%%%%%%%
\label{sec:stage2}

% \subsubsection{View-Aware Texture Prediction via Tweedie's Formula}
\vspace{-.5em}
\subsubsection{View-Aware Texture Prediction.}
Because the fused Stage~1 mesh contains unnatural geometry, direct TRELLIS texturing fails. Therefore, we treat texturing as an independent stage using a depth-conditioned ControlNet~\cite{zhang2023adding} (Stable Diffusion~\cite{rombach2022high}) to predict clean images $\hat{x}_{1|t}$. At each denoising step, we render the mesh from $\theta_1$ and $\theta_2$, predict textures, and un-project them onto the 3D surface. Finally, Mesh Texture Aggregation merges these multi-view contributions using cosine-weighted blending (based on surface normals), ensuring high-quality, seamless textures.

\vspace{-.5em}
\subsubsection{View-Dependent Texture Selection.}
We determine which branch's texture to use based on the current camera angle. Taking $\theta_1 = 0^\circ$ as an example, viewpoints within $270^\circ$--$90^\circ$ adopt the texture estimate from $y_1$, while the remaining angles adopt that from $y_2$. Although switched via a hard cutoff, no visible seam appears in practice, as cosine-weighted blending naturally smooths the transition at the boundary.

% To ensure semantic consistency of the texture at each viewpoint, we determine which branch's texture prediction to use based on the current camera angle. Concretely, taking $\theta_1 = 0^\circ$ as an example, viewpoints within the range $270^\circ$--$90^\circ$ adopt the texture estimate from $y_1$, while the remaining angles adopt the texture estimate from $y_2$, allowing each facing direction to present the corresponding semantic texture details.

% Although the two semantics are switched via a hard cutoff at the boundary angle, no visible seam appears in practice. This is because the cosine-weighted averaging already provides spatial smoothing: surface points near the boundary are typically covered by both viewpoints simultaneously, and the texture contributions from the two branches are naturally blended through cosine weighting, producing a smooth semantic transition at the boundary rather than an abrupt switch.

%%%%%%%%%%%%%%%%%%%%%%%%%%%%%%%%%%%%%%%%%%%%%%%%%%
\vspace{-.5em}
\subsection{Noise Guidance}
\vspace{-.5em}
%%%%%%%%%%%%%%%%%%%%%%%%%%%%%%%%%%%%%%%%%%%%%%%%%%

\begin{figure*}[t]
\vspace{-2mm}
  \centering
  \includegraphics[width=0.8\textwidth]{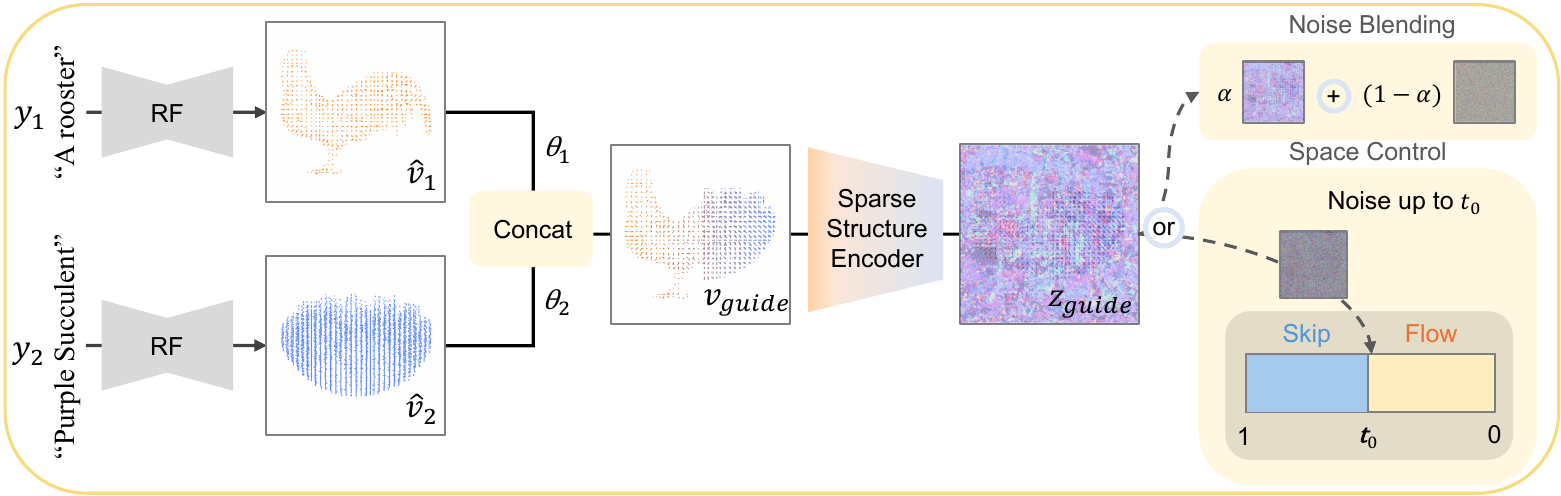}
  % \caption{
  % \textbf{Noise Guidance.}
  % Given two prompts $y_1$ and $y_2$, two single-semantic voxels $\hat{v}_1$ and $\hat{v}_2$ are independently generated from the same noise and concatenated according to target angles $\theta_1$ and $\theta_2$ to form a guidance voxel $v_\text{guide}$, which is then encoded into a guidance latent $z_\text{guide}$ via the Sparse VAE Encoder.
  % \textbf{Noise Blending Guidance} mixes $z_\text{guide}$ with pure noise as $\alpha \cdot z_\text{guide} + (1-\alpha) \cdot z_\text{noise}$, injecting a mild spatial prior while preserving generation diversity ($\alpha = 0.3$ in our experiments).
  % \textbf{Space Control Guidance} instead controls the starting denoising timestep $t_0$, interpolating between the guided latent and pure noise as $t_0 \cdot \text{Encoder}(v_\text{guide}) + (1 - t_0) \cdot z_\text{noise}$; the latent state is fully guided for the first $t_0$ steps before normal denoising resumes, providing stronger structural constraints for object pairs with large geometric discrepancies ($t_0 = 10$ in our experiments).
  % }
  \vspace{-3mm}
  \caption{
  \textbf{Noise Guidance.}
  Given two prompts $y_1$ and $y_2$, two single-semantic voxels $\hat{v}_1$ and $\hat{v}_2$ are independently generated and concatenated at $\theta_1$ and $\theta_2$ to form $v_\text{guide}$, encoded into $z_\text{guide}$ via the Sparse Structure Encoder.
  \textbf{Noise Blending Guidance} mixes $z_\text{guide}$ with pure noise via $\alpha \cdot z_\text{guide} + (1-\alpha) \cdot z_\text{noise}$, injecting a mild spatial prior.
  \textbf{Space Control Guidance} interpolates between $z_\text{guide}$ and noise at timestep $t_0$, providing stronger structural constraints for geometrically challenging pairs.
  }
  \vspace{-2mm}
  \label{fig:noise_guidance}
  \vspace{-3mm}
\end{figure*}

% The starting point of Rectified Flow denoising is pure random noise, which for dual-semantic fusion lacks any spatial structural constraint, making it prone to geometric interference between the two semantics and difficult convergence. To address this, we propose two noise guidance strategies that inject a spatial prior before denoising begins, improving the quality of geometric fusion. The two strategies are illustrated in Fig.~\ref{fig:noise_guidance}.

The starting point of Rectified Flow denoising---pure random noise---lacks spatial structural constraint for dual-semantic fusion, making it prone to geometric interference and difficult convergence. We propose two noise guidance strategies that inject a spatial prior before denoising begins, improving geometric fusion quality, as illustrated in Fig.~\ref{fig:noise_guidance}.

\vspace{-.5em}
\subsubsection{Noise Blending Guidance.}
% We independently denoise from the same noise using $y_1$ and $y_2$ to generate two single-semantic voxels $\hat{v}_1$ and $\hat{v}_2$. The two voxels are each halved and concatenated according to target angles $\theta_1$ and $\theta_2$ to form a guidance voxel $v_\text{guide}$ that encodes a spatial distribution prior. Gaussian noise is then added to $v_\text{guide}$, which is encoded into a guidance latent and mixed with pure noise in a weighted manner:
We pre-generate two single-semantic voxels $\hat{v}_1$ and $\hat{v}_2$, halve and concatenate them at target angles $\theta_1$ and $\theta_2$ to form a guidance voxel $v_\text{guide}$, which is encoded into a guidance latent via the Sparse Structure Encoder. The guidance latent is then combined with pure Gaussian noise via a weighted sum as the initial denoising latent:
\begin{equation}\small
    z_\text{init} = \alpha \cdot \text{Encoder}(v_\text{guide}) + (1 - \alpha) \cdot z_\text{noise},
\end{equation}
where $\alpha$ controls the guidance strength, balancing structural prior and generation diversity.
% where the hyperparameter $\alpha$ controls the guidance strength. In our experiments, we set $\alpha = 0.3$, allowing the generated result to retain the structural prior from the guidance while preserving sufficient randomness to maintain generation diversity.

\vspace{-.5em}
\subsubsection{Space Control Guidance.}
% Inspired by SpaceControl~\cite{fedele2025spacecontrol}, which introduces spatial control into 3D generative models by manipulating the starting denoising timestep, we adapt this principle to our dual-semantic fusion setting.

% Similarly, we first generate the guidance voxel $v_\text{guide}$ and encode it into a guidance latent $z_\text{guide}$. Unlike Noise Blending Guidance, this strategy controls the starting denoising timestep $t_0$ to interpolate between the guidance signal and noise in latent space:

Inspired by SpaceControl~\cite{fedele2025spacecontrol}, we adapt the principle of manipulating the starting denoising timestep to our dual-semantic fusion setting. Similarly to Noise Blending Guidance, we first generate $v_\text{guide}$ and encode it into a guidance latent $z_\text{guide}$, and interpolate between the guidance latent and pure noise at timestep $t_0$:
\begin{equation}\small
    z\{t_0\} = t_0 \cdot \text{Encoder}(v_\text{guide}) + (1 - t_0) \cdot z_\text{noise}.
\end{equation}
A larger $t_0$ imposes stronger structural constraints, while a smaller $t_0$ preserves more generation diversity. In our 25-step setting, $t_0 = 10$ means the first 10 steps are guided before normal denoising resumes, striking a balance between structural constraint and generation freedom.
% A larger $t_0$ brings the initial latent estimate closer to the structure of the guidance voxel, while a smaller $t_0$ preserves more randomness to increase generation diversity. In our 25-step denoising setting, we set $t_0 = 10$, meaning the first 10 steps operate under the guided latent state before normal denoising iterations resume, striking a balance between structural constraint and generation freedom.

%%%%%%%%%%%%%%%%%%%%%%%%%%%%%%%%%%%%%%%%%%%%%%%%%%
\vspace{-.5em}
\subsection{CLIP-guided Orientation Search}
\vspace{-.5em}
%%%%%%%%%%%%%%%%%%%%%%%%%%%%%%%%%%%%%%%%%%%%%%%%%%

\begin{figure*}[t]
\vspace{-3mm}
  \centering
  \includegraphics[width=0.65\textwidth]{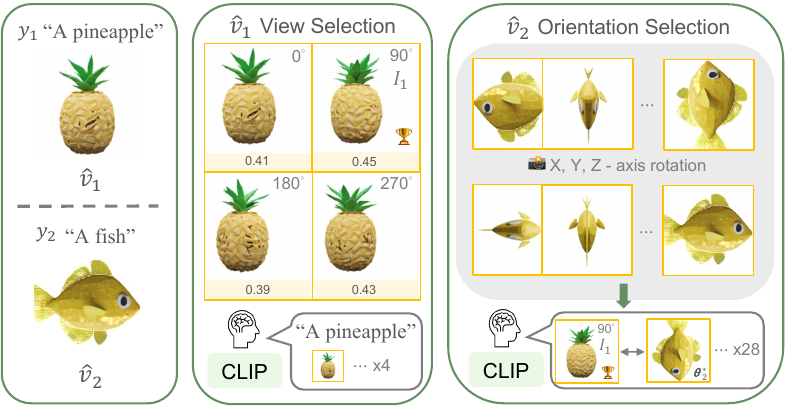}
  % \caption{
  % \textbf{CLIP-guided Orientation Search.}
  % Given two single-semantic voxels $\hat{v}_1$ and $\hat{v}_2$,
  % \textbf{Anchor View Selection} renders 4 candidate views of $\hat{v}_1$ at $90^\circ$ intervals around the $z$-axis and selects the view with the highest CLIP text-image similarity to prompt $y_1$ as the representative view $I_1$ (e.g., $90^\circ$ with score $0.45$ for \textit{``A pineapple''}).
  % \textbf{Cross-object Orientation Matching} generates 28 candidate renders of $\hat{v}_2$ by sampling discrete rotation combinations along the $(X, Y, Z)$ axes, computes the CLIP image-image similarity between each render and $I_1$, and selects the rotation with the highest similarity as the fusion orientation $\theta_2^*$ for object~2.
  % This ensures that the geometric silhouettes of the two objects are maximally aligned prior to SDF blending, replacing the default fixed-angle configuration in the subsequent dual-branch denoising process.
  % }
  \vspace{-3mm}
  \caption{
  \textbf{CLIP-guided Orientation Search.} \textbf{Anchor View Selection} identifies the best representative view $I_1$ for $\hat{v}_1$ from 4 orthogonal renders via CLIP text-image similarity (e.g., \textit{``pineapple''} at $90^\circ$). \textbf{Cross-object Matching} then evaluates 28 3D rotations of $\hat{v}_2$, selecting the angle $\theta_2^*$ that maximizes CLIP image-image similarity to $I_1$. This optimally aligns their silhouettes prior to SDF blending.
  % \textbf{CLIP-guided Orientation Search.}
  % \textbf{Anchor View Selection} renders 4 candidate views of $\hat{v}_1$ at $90^\circ$ intervals and selects the highest CLIP text-image similarity as representative view $I_1$ (e.g., $90^\circ$, score $0.45$ for \textit{``A pineapple''}).
  % \textbf{Cross-object Orientation Matching} generates 28 candidate renders of $\hat{v}_2$ across $(X, Y, Z)$ axes and selects the rotation $\theta_2^*$ with the highest CLIP image-image similarity to $I_1$, maximally aligning the two silhouettes prior to SDF blending.
  }
  \vspace{-2mm}
  \label{fig:clip_rotation}
  \vspace{-3mm}
\end{figure*}

% Existing multi-view illusion methods typically assume fixed viewpoints (e.g., $0^\circ$ and $180^\circ$); however, different objects may exhibit significant orientation differences in their canonical poses as generated by diffusion models. Directly fusing two objects with substantially different geometric orientations at the default angles often results in a blended mesh that fails to present clear semantic silhouettes from any viewpoint. To address this, we propose a CLIP-similarity-based adaptive orientation search mechanism that automatically selects the optimal fusion orientation for both objects before denoising begins. The search procedure is illustrated in Fig.~\ref{fig:clip_rotation}.
% In all experiments, we use the ViT-B/32 model from OpenCLIP~\cite{ilharco_gabriel_2021_5143773}, pretrained on LAION-2B with weights \texttt{laion2b\_s34b\_b79k}, to compute similarity scores.

Existing multi-view illusion methods typically assume fixed viewpoints (e.g., $0^\circ$ and $180^\circ$); however, different objects may exhibit significant orientation differences in their canonical poses. Directly fusing two geometrically misaligned objects at default angles often results in a blended mesh that fails to present clear semantic silhouettes. To address this, we propose a CLIP-based adaptive orientation search that automatically selects the optimal fusion orientation before denoising begins, as illustrated in Fig.~\ref{fig:clip_rotation}.

\vspace{-.5em}
\subsubsection{Anchor View Selection for Object~1.}
% We independently generate single-semantic voxels $\hat{v}_1$ and $\hat{v}_2$ using $y_1$ and $y_2$ respectively. For object~1, we fix it at its canonical orientation and render 4 candidate views at $90^\circ$ intervals around the $Z$-axis. The view with the highest CLIP text-image similarity to prompt $y_1$ is selected as the representative view $I_1$ of object~1.
We independently generate single-semantic voxels $\hat{v}_1$ and $\hat{v}_2$ from prompts $y_1$ and $y_2$ respectively. For object~1, we render 4 candidate views of $\hat{v}_1$ at $90^\circ$ intervals around the $Z$-axis and select the view with the highest CLIP text-image similarity to $y_1$ as the representative view $I_1$.

\vspace{-.5em}
\subsubsection{Cross-Object Orientation Matching for Object~2.}
For object~2, we sample discrete rotation combinations along the $(X, Y, Z)$ axes, generating 28 candidate renders, and select the rotation with the highest CLIP image-image similarity to $I_1$ as the fusion orientation $\theta_2^*$. Candidate angles are sampled at $90^\circ$ intervals, covering $\{0^\circ, 90^\circ, 180^\circ, 270^\circ\}$ per axis, balancing search efficiency and angular coverage. Denser sampling (e.g., $45^\circ$) could further improve alignment precision, which we leave as future work.

% For object~2, we sample discrete rotation combinations along the $(X, Y, Z)$ axes, generating a total of 28 candidate renders. The CLIP image-image similarity between each candidate render and the representative view $I_1$ is computed, and the rotation combination with the highest similarity is selected as the fusion orientation $\theta_2^*$ for object~2. This ensures that the geometric silhouettes of the two objects are maximally aligned, and $\theta_2^*$ replaces the default fixed angle for voxel rotation and blending in the subsequent dual-branch denoising process.

% In our experiments, candidate angles are sampled at $90^\circ$ intervals, covering all combinations of $\{0^\circ, 90^\circ, 180^\circ, 270^\circ\}$ per axis. This setting balances search efficiency and angular coverage, and is sufficient to handle the orientation discrepancies of most object pairs. The method can naturally be extended to denser angular sampling (e.g., $45^\circ$ or finer intervals) for improved geometric alignment precision, which we leave as future work.

%%%%%%%%%%%%%%%%%%%%%%%%%%%%%%%%%%%%%%%%%%%%%%%%%%
\vspace{-.5em}
\subsection{Extension to Three-Object 3D Illusions}
\label{sec:three_object}
\vspace{-.5em}
%%%%%%%%%%%%%%%%%%%%%%%%%%%%%%%%%%%%%%%%%%%%%%%%%%

\begin{figure*}[t]
\vspace{-2mm}
  \centering
  \includegraphics[width=0.9\textwidth]{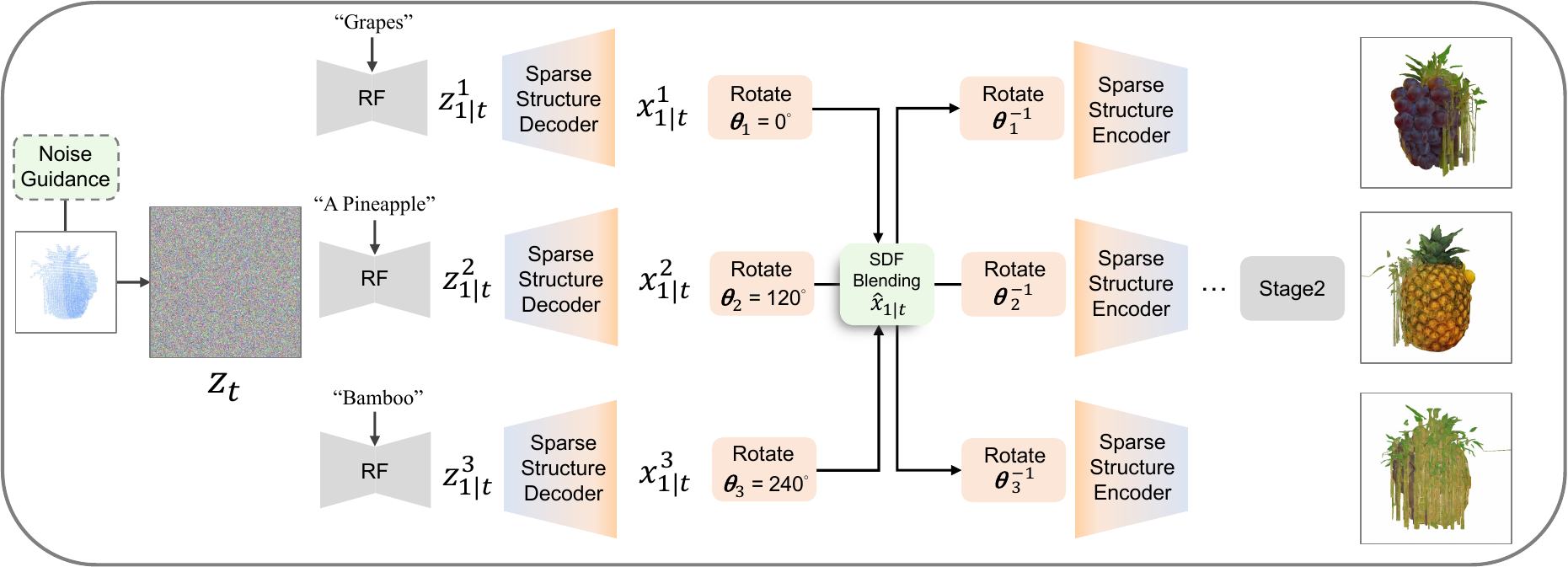}
  % \caption{
  % \textbf{Extension to three-object 3D illusion.}
  % Our framework naturally extends to three-object illusion generation by adding a third denoising branch, without modifying the fusion procedure or optimization pipeline.
  % Three branches are conditioned on prompts $y_1$ (\textit{``Grapes''}), $y_2$ (\textit{``A Pineapple''}), and $y_3$ (\textit{``Bamboo''}), all starting from the same noise $z_t$.
  % The rotation angles are fixed at $\theta_1 = 0^\circ$, $\theta_2 = 120^\circ$, and $\theta_3 = 240^\circ$, uniformly distributing the three target viewpoints across $360^\circ$.
  % At each timestep, the three predicted clean voxels are rotation-aligned, SDF-fused into a blended voxel $\hat{x}_{1|t}$, inverse-rotated, and re-encoded into each branch's latent space for the next denoising step.
  % Noise Guidance is required in this setting to provide a stable structural prior that steers the three semantic branches toward their respective targets.
  % The Stage~2 view-conditioned texture synthesis proceeds identically to the two-object case, generating semantically consistent textures from all three target viewpoints.
  % }
\vspace{-2mm}
  \caption{
  \textbf{Extension to three-object 3D illusion.}
  Our framework scales to three semantics (e.g., \textit{``Grapes''}, \textit{``A Pineapple''}, \textit{``Bamboo''}) by adding a third denoising branch, with rotation angles fixed at $0^\circ$, $120^\circ$, $240^\circ$ to uniformly cover $360^\circ$. Noise Guidance steers all three branches toward their respective targets, ensuring each semantic is clearly presented at its target viewpoint.
  }
  \vspace{-2mm}
  \label{fig:three_phase}
\vspace{-3mm}
\end{figure*}

Our framework naturally scales to three-object illusions by adding a third Stage~1 denoising branch. Given three prompts and a shared noise $z_t$, we fix target angles at $0^\circ$, $120^\circ$, and $240^\circ$. At each step, the three predicted voxels are SDF-fused, inverse-rotated, and re-encoded. Due to complex geometric conflicts, Noise Guidance is mandatory here, utilizing a merged prior of three pre-generated single-semantic objects. {Specifically, since averaging three geometries imposes stronger geometric conflicts than the two-object case, we adopt Space Control Guidance with an increased $t_0 = 20$ out of 25 steps, providing stronger structural constraints to ensure each semantic is preserved at its target viewpoint.} Stage~2 remains identical to the two-object case. See Fig.~7 for the pipeline and Sec.~4.4 for results.

% Our framework naturally extends to three-object illusion generation by adding a third denoising branch to Stage~1, without modifying the fusion procedure. Given three text prompts, the branches start from the same noise $z_t$ with rotation angles fixed at $0^\circ$, $120^\circ$, $240^\circ$, uniformly distributing the three semantics across $360^\circ$. At each timestep, the three predicted voxels are SDF-fused, inverse-rotated, and re-encoded for the next denoising step. Noise Guidance is required due to increased geometric conflicts; we pre-generate three single-semantic objects at the target angles and combine them as an initial guidance prior. Stage~2 proceeds identically to the two-object case. The pipeline is illustrated in Fig.~\ref{fig:three_phase} and qualitative results are presented in Sec.~\ref{sec:applications}.

\vspace{-.5em}
\section{Experiments}

\vspace{-.5em}
\subsection{Experimental Setup}

% \noindent\textbf{Baselines.}
% Since prior work directly addressing the same task remains scarce, we compare against two baselines. The first is \textbf{Shape from Semantics}~\cite{li2025shapesemantics3dshape}, which tackles the identical task setting via Score Distillation Sampling (SDS) optimization, but requires approximately 40 minutes per generated object, incurring substantially higher computational cost than our method. The second is \textbf{Direct Concatenation}, which independently generates the geometry of two objects using TRELLIS\cite{xiang2025structured}, then naively halves each along the midplane and stitches them into a single mesh, serving as the simplest possible fusion strategy.

% origin baseline section
% \vspace{-.5em}
% \subsubsection{Baselines.}
% We compare against two baselines: \textbf{Shape from Semantics}~\cite{li2025shapesemantics3dshape}, which addresses the same task via Score Distillation Sampling (SDS) but requires $\sim$40 minutes per object; and \textbf{Direct Concatenation}, which independently generates two objects via TRELLIS~\cite{xiang2025structured}, halves each along the midplane, and stitches them into a single mesh.

\vspace{-.5em}
\subsubsection{Baselines.}
We compare against \textbf{five} baselines: 
\textbf{Shape from Semantics}~\cite{li2025shapesemantics3dshape}, which addresses 
the same task via Score Distillation Sampling (SDS) but requires $\sim$40 minutes 
per object; \textbf{Shape from Semantics (VSD)}~\cite{li2025shapesemantics3dshape}, 
a stronger variant that replaces the SDS objective with Variational Score Distillation 
(VSD)~\cite{wang2023prolificdreamer}, incurring $\sim$4 hr/sample runtime; 
and \textbf{Direct Concatenation}, which independently generates two objects 
via TRELLIS~\cite{xiang2025structured}, halves each along the midplane, and stitches 
them into a single mesh. We also evaluate \textbf{TRELLIS}~\cite{xiang2025structured} 
and \textbf{DreamBeast}~\cite{li2025dreambeast}, both prompted with \textit{``a single 
3D object, front side is \{$y_1$\}, back side is \{$y_2$\}''} to elicit view-dependent 
semantics from a single generation. TRELLIS is a state-of-the-art feed-forward 3D 
generation model, while DreamBeast is an SDS-based method designed for generating 
fantastical creatures with part-level semantic control. These two baselines assess 
whether existing generation methods can produce coherent visual illusions without 
task-specific design.

\vspace{-.5em}
\subsubsection{Data.}
We collect object prompts from Shape from Semantics~\cite{li2025shapesemantics3dshape} and supplement with additional common objects, yielding 60 distinct objects across five categories: 16 birds, 19 mammals, 5 reptiles and aquatic animals, 9 plants, and 11 man-made artifacts. Two objects are randomly sampled per experiment to form a prompt pair.

\vspace{-.5em}
\subsubsection{Implementation Details.}
All experiments are conducted on a single NVIDIA RTX 4090 GPU. For SDF blending, we use Truncated SDFs with truncation distance $\text{clip\_s}=12$ and binarization threshold $\tau=0.8$. Stage~1 runs for 25 denoising steps with one geometry blending operation per step, with CFG applied at guidance scale $\omega=7.5$ within $t\in[0.5,0.95]$ (Interval CFG); Stage~2 runs for 30 denoising steps. For CLIP-guided Orientation Search, we use OpenCLIP~\cite{ilharco_gabriel_2021_5143773} ViT-B/32 pretrained on LAION-2B (\texttt{laion2b\_s34b\_b79k}). For Noise Blending Guidance, we set $\alpha=0.3$; for Space Control Guidance, we set $t_0=10$ out of 25 steps.

We organize experiments into three rotational configurations, with object~$A$ fixed at its canonical orientation: \textbf{Case~1}, object~$B$ is unrotated, revealing the front of object~$A$ at $0^\circ$ and the back of object~$B$ at $180^\circ$; \textbf{Case~2}, object~$B$ is rotated by $180^\circ$, revealing the front views of both objects at the two target viewpoints; and \textbf{Case~3}, object~$B$ is automatically rotated via CLIP-guided Orientation Search, typically placing the two objects approximately $180^\circ$ apart.

Noise Blending and Space Control Guidance can be freely combined with Cases~1 and~2, whereas Case~3 requires no additional guidance since the fusion angle is already adapted via CLIP. Cases~1 and~2 require approximately 3 minutes in total ($\sim$1 min for Stage~1 and $\sim$2 min for Stage~2), while Case~3 requires approximately 5 minutes due to the additional CLIP-guided Orientation Search ($\sim$3 min for Stage~1 and $\sim$2 min for Stage~2). All configurations complete within 3--5 minutes, offering a significant efficiency advantage over the SDS-based baseline ($\sim$40 minutes).

\vspace{-.5em}
\subsubsection{Metrics.}
\label{sec:metrics}
We design six quantitative metrics and a user study to evaluate generation quality and illusion effect: 
% \textit{(1) CLIP Similarity.}
% We compute CLIP~\cite{radford2021learning, ilharco_gabriel_2021_5143773} text-image similarity across 1,000 renders per viewpoint ($\pm20^\circ$ jitter). Notably, Direct Concatenation artificially inflates this score by trivially preserving single-view appearances via naive stitching, making CLIP alone insufficient.
% \textit{(2) GPT Accuracy (\%).}
% To evaluate semantic clarity, we prompt GPT-4.1-mini to identify the respective semantics from side-by-side renders of both target viewpoints. {Specifically, given a side-by-side image composed of two viewpoint renders (labeled A and B), we query the model with: \textit{``Given the image with pieces A and B, choose which interpretation fits: Option 1: Left = \{$y_1$\}, Right = \{$y_2$\}; Option 2: Left = \{$y_2$\}, Right = \{$y_1$\}. Respond with Option 1 or Option 2 only.''} Accuracy is computed as the proportion of responses matching the ground-truth prompt assignment.} This 2-way accuracy quantifies how well the object conveys its intended semantics.

\textit{(1) CLIP Similarity} computes CLIP~\cite{radford2021learning, ilharco_gabriel_2021_5143773} text-image similarity across 1,000 renders per viewpoint ($\pm20^\circ$ jitter). Note that Direct Concatenation artificially inflates this score by trivially preserving single-view appearances, making CLIP alone insufficient.

\begin{wrapfigure}{r}{0.42\columnwidth}
  \vspace{-6mm}
  \centering
  \includegraphics[width=0.42\columnwidth]{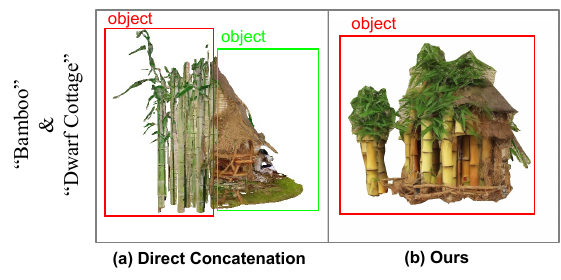}
  \vspace{-3mm}
  \caption{
  \textbf{Object detection at the junction viewpoint} for \textit{``Bamboo''} \& \textit{``Dwarf Cottage''}.
  (a)~Direct Concatenation is detected as two separate objects (red and green boxes);
  (b)~Ours is detected as a single unified object (red box only).
  }
  \label{fig:object_detection}
  \vspace{-4mm}
\end{wrapfigure}

\textit{(2) GPT Accuracy (\%)} prompts GPT-4.1-mini to identify target semantics from side-by-side renders of both viewpoints, reporting the proportion of responses matching the ground-truth prompt assignment as a 2-way accuracy.

\textit{(3) FID \& KID} measure visual realism between 1{,}000 renders and 1{,}000 reference images (20 views of 50 objects) from Objaverse~1.0~\cite{deitke2023objaverse}.
\textit{(4) Object Detection Score} renders the midpoint angle and counts objects via OWLv2~\cite{minderer2023scaling} (Fig.~\ref{fig:object_detection}), reporting average object count (ideally 1) and multi-object rate (proportion with $>1$ detection).
\textit{(5) View-Conditional CLIP Contrast} computes CLIP similarity between each viewpoint render and the \emph{opposite} prompt, penalizing cross-view semantic leakage that naive stitching permits; a lower score reflects cleaner perceptual separation.
\textit{(6) Boundary Seam Score} counts triangles connected to fewer than three neighbors (deg=1, lower is better), averaged over 50 pairs; Direct Concatenation yields 1503.42 vs.\ \textbf{798.31} for ours (Fig.~\ref{fig:qualitative_normal_map}).
\vspace{-4mm}

\vspace{-.5em}
\subsubsection{User Study.}
Beyond quantitative metrics, we conduct a user study to collect perceptual evaluations from human observers.
{A total of 50 participants took part in the study. Each participant was presented with rendered results from all three methods and asked to evaluate them across the following questions, each with four prompt-pair sub-questions:}
% Participants are asked to answer the following questions:
\begin{itemize}
    \item \textbf{Q1:} How recognizable are the intended semantics at each target viewpoint? (1: unrecognizable, 2: partially recognizable, 3: clearly recognizable)
    \item \textbf{Q2:} Which result better aligns with the intended semantics?
    \item \textbf{Q3:} Comparing CLIP-adaptive orientation (Case~3) versus fixed $0^\circ$/$180^\circ$ angles, which produces a more natural illusion effect?
    \vspace{-6mm}
\end{itemize}

%%%%%%%%%%%%%%%%%%%%%%%%%%%%%%%%%%%%%%%%%%%%%%%%%%
\vspace{-.5em}
\subsection{Results and Analysis}

\begin{figure*}[t]
\vspace{-1mm}
  \centering
  \includegraphics[width=\textwidth]{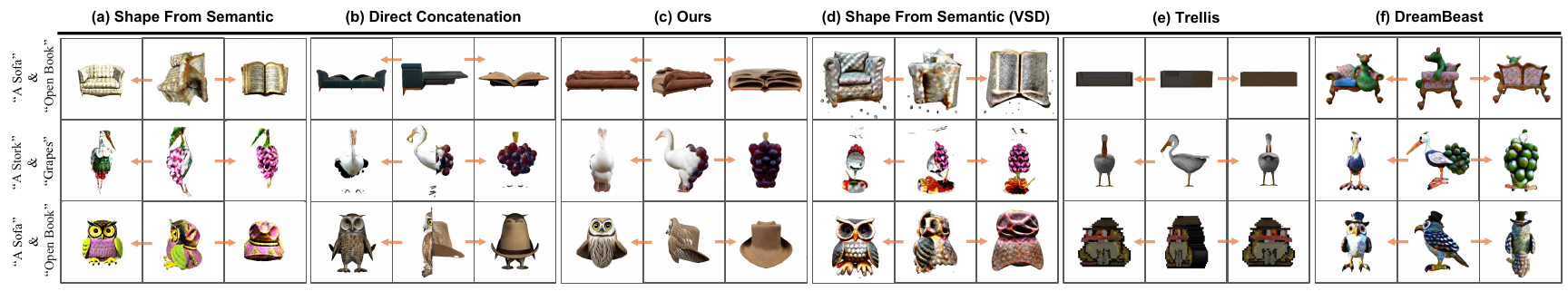}
\vspace{-6mm}
\caption{
  \textbf{Qualitative comparison with baselines.} Left-to-right: View~1, blended mesh, View~2.
  \textbf{(a)}~Shape From Semantics~\cite{li2025shapesemantics3dshape} suffers from over-saturation and geometry leakage.
  \textbf{(b)}~Direct Concatenation exposes unnatural junction seams and opposing geometry at target views.
  \textbf{(c)}~Ours produces a coherent mesh with clear, view-dependent semantics and no leakage.
  \textbf{(d)}~Shape From Semantics (VSD)~\cite{li2025shapesemantics3dshape} still exhibits over-saturation and leakage despite its stronger objective.
 \textbf{(e)}~TRELLIS~\cite{xiang2025structured}, prompted with \textit{``a single 3D object, front side is A, back side is B''}, fails to produce view-dependent semantics and collapses to a generic shape that reflects neither target interpretation. 
  \textbf{(f)}~DreamBeast~\cite{li2025dreambeast} blends both semantics uniformly across all viewpoints, failing to achieve view-dependent separation.
  }
  % \caption{
  % \textbf{Qualitative comparison with baselines.} Left-to-right: View~1, blended mesh, View~2. 
  % \textbf{(a)}~Shape From Semantics~\cite{li2025shapesemantics3dshape} suffers from over-saturation and geometry leakage. 
  % \textbf{(b)}~Direct Concatenation exposes unnatural junction seams and the opposing geometry at target views. 
  % \textbf{(c)}~Ours produces a single coherent mesh with clear, view-dependent semantics and no leakage. 
  % \textbf{(d)}~Shape From Semantics (VSD)~\cite{li2025shapesemantics3dshape}, despite employing a stronger optimization objective, still exhibits color over-saturation and semantic leakage artifacts. 
  % \textbf{(e)}~TRELLIS~\cite{xiang2025structured}, prompted with \textit{``a single 3D object, front side is A, back side is B''}, fails to produce view-dependent semantics and collapses to a generic shape that reflects neither target interpretation. 
  % \textbf{(f)}~DreamBeast~\cite{li2025dreambeast}, despite its part-level semantic control, generates fantastical hybrid creatures that blend both semantics into every viewpoint rather than isolating them, failing to achieve the intended illusion effect.
  % }
  \label{fig:qualitative_fixed}
\vspace{-3mm}
\end{figure*}

\begin{figure*}[t]
\vspace{-1mm}
  \centering
  \includegraphics[width=\textwidth]{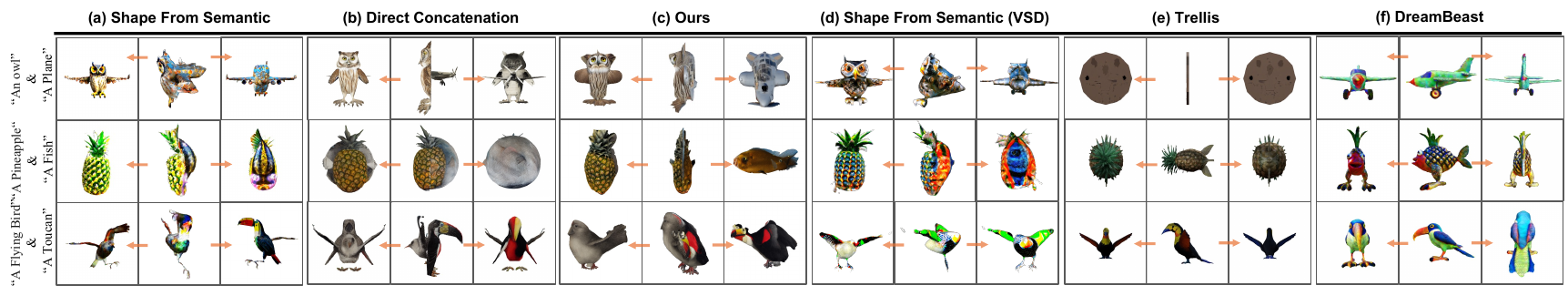}
\vspace{-6mm}
\caption{
  \textbf{Qualitative comparison (CLIP-guided orientation).} Left-to-right: View 1, blended mesh, View 2.
  \textbf{(c) Ours} uses CLIP-guided search to adaptively align compatible silhouettes, revealing clearer semantics at both viewpoints.
  \textbf{(a)(b)} Shape From Semantics~\cite{li2025shapesemantics3dshape} and Direct Concatenation suffer from silhouette misalignment and degraded fusion without adaptive rotation.
  \textbf{(d)(e)(f)} See Fig.~\ref{fig:qualitative_fixed} caption for descriptions of remaining baselines.
  }
  \vspace{-2mm}
  % \caption{
  % \textbf{Qualitative comparison (CLIP-guided orientation).} Left-to-right: View 1, blended mesh, View 2. 
  % \textbf{(c) Ours} uses CLIP-guided search to adaptively align compatible silhouettes (e.g., sideways fish with pineapple, flying owl with plane), revealing clearer semantics at both viewpoints. 
  % Lacking adaptive rotation, \textbf{(a)}~Shape From Semantics~\cite{li2025shapesemantics3dshape} and \textbf{(b)}~Direct Concatenation suffer from silhouette misalignment and degraded fusion. 
  % \textbf{(d)}~Shape From Semantics (VSD)~\cite{li2025shapesemantics3dshape}, despite its stronger optimization objective, similarly fails to recover clear semantics under silhouette misalignment. 
  % \textbf{(e)}~TRELLIS~\cite{xiang2025structured}, prompted with \textit{``a single 3D object, front side is A, back side is B''}, produces flat, degenerate geometry that fails to capture either semantic. 
  % \textbf{(f)}~DreamBeast~\cite{li2025dreambeast} blends both semantics uniformly across all viewpoints, generating fantastical hybrid creatures rather than achieving view-dependent separation.
  % }
  \label{fig:qualitative_clip}
\vspace{-3mm}
\end{figure*}

\vspace{-.5em}
\subsubsection{Quantitative Comparison.}

\begin{wrapfigure}{r}{0.48\columnwidth}
\vspace{-4mm}
\centering
\includegraphics[width=0.48\columnwidth]{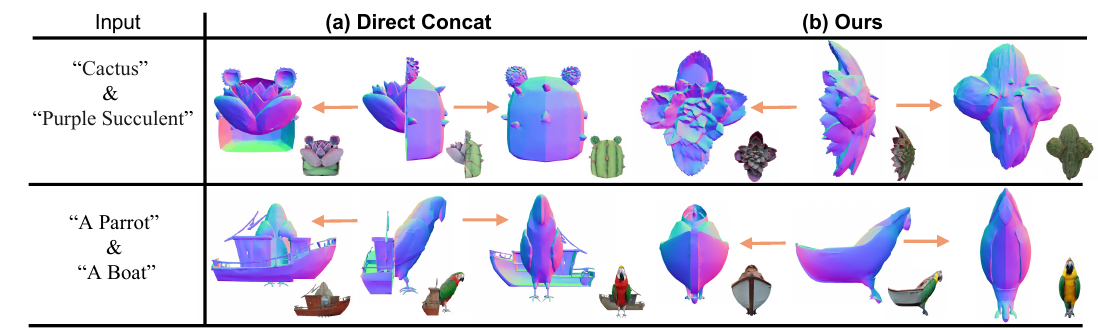}
\vspace{-4mm}
\caption{
\textbf{Normal map comparison at fusion boundary.}
Direct Concatenation exhibits sharp discontinuities, whereas ours produces smoother normals across the fusion region.
}
\label{fig:qualitative_normal_map}
\vspace{-6mm}
\end{wrapfigure}

Table~\ref{tab:comparison} reports results over 50 randomly sampled prompt pairs.
Although Direct Concatenation achieves the highest CLIP score (29.030 vs.\ ours 28.170), this is a systematic artifact of naive per-viewpoint appearance preservation rather than genuine illusion quality (Sec.~\ref{sec:metrics}).
Our method achieves the highest GPT Accuracy (84\%), lowest FID (185.555), and best Object Detection scores (avg.\ count 0.86, multi-object rate 18\% vs.\ 56\% for Direct Concatenation), demonstrating superior semantic recognizability, visual realism, and geometric coherence.
Table~\ref{tab:comparison} further reports a lower Boundary Seam Score for ours (\textbf{798.31} vs.\ 1503.42 for Direct Concatenation), while Fig.~\ref{fig:qualitative_normal_map} visually confirms smoother normals at the fusion boundary.
All configurations complete within 3--5 minutes, matching Direct Concatenation and offering a significant efficiency advantage over Shape from Semantics~\cite{li2025shapesemantics3dshape} ($\sim$40 minutes).
\vspace{-4mm}

% Table~\ref{tab:comparison} reports results over 50 randomly sampled prompt pairs. Our method outperforms or matches both baselines on the majority of metrics. Direct Concatenation achieves the highest CLIP score (29.030 vs.\ ours 28.170), but as discussed in Sec.~\ref{sec:metrics}, this is a systematic artifact of naively preserving per-viewpoint appearance rather than genuine illusion quality.
% Our method achieves the highest GPT Accuracy (84\%), outperforming Direct Concatenation (76\%) and Shape from Semantics (70\%), lowest FID (185.555), and best Object Detection scores (avg.\ count 0.86, multi-object rate 18\% vs.\ 2.1 and 56\% for Direct Concatenation), collectively demonstrating superior semantic recognizability, visual realism, and geometric coherence. Our method completes within 3--5 minutes, matching Direct Concatenation and offering a significant efficiency advantage over Shape from Semantics~\cite{li2025shapesemantics3dshape} ($\sim$40 minutes).

% Normal map renderings further confirm the geometric advantage of our method: 
% our fusion boundaries exhibit smooth surface normals, whereas Direct 
% Concatenation produces sharp discontinuities at the stitching seam 
% (Fig.~\ref{fig:qualitative_normal_map}), quantified by a boundary seam score of 
% \textbf{798.31} vs.\ 1503.42 (deg=1, lower is better).

\vspace{-.5em}
\subsubsection{Qualitative Comparison.}
Fig.~\ref{fig:qualitative_fixed} and Fig.~\ref{fig:qualitative_clip} compare results under fixed-angle and CLIP-guided configurations respectively. Our method produces geometrically coherent results with clear per-viewpoint semantics, while Shape From Semantics~\cite{li2025shapesemantics3dshape} suffers from over-saturation and Direct Concatenation exposes visible junction seams.
\vspace{-4mm}

\begin{table}[t]
\centering
\setlength{\tabcolsep}{4pt}
\caption{\textbf{Comparison with the baselines.}}
\vspace{-3mm}
\resizebox{\columnwidth}{!}{%
\begin{tabular}{lcccccccccc}
\toprule
\multirow{2}{*}{Method} & \multirow{2}{*}{CLIP $\uparrow$} & \multirow{2}{*}{CLIP (opp.) $\downarrow$} & \multirow{2}{*}{GPT Acc. (\%) $\uparrow$} & \multirow{2}{*}{FID $\downarrow$} & \multirow{2}{*}{KID $\downarrow$} & \multicolumn{2}{c}{Object Detection} & \multirow{2}{*}{Boundary Seam (deg=1) $\downarrow$} & \multirow{2}{*}{Runtime} \\
\cmidrule(lr){7-8}
 & & & & & & Avg. Obj. Count & Multi-Obj. Rate (\%) & & \\
\midrule
Shape From Semantic~\cite{li2025shapesemantics3dshape}       & 27.460          & 19.72          & 70          & 194.136          & 0.051          & 0.64 & 2  & -- & $\sim$40 min   \\
Shape From Semantic (VSD)~\cite{li2025shapesemantics3dshape} & 27.523          & 20.69             & 65          & 195.160          & 0.061          & 0.68 & 4  & --    & $\sim$3-4 hr   \\
Direct Concat                                                & \textbf{29.030} & 20.38          & 76          & 187.886          & 0.067          & 2.1  & 56 & 1503.42 & $\sim$3-5 min  \\
Trellis                                                      & 22.180          & 22.68          & 60          & \textbf{174.130} & \textbf{0.044} & 0.58 & 8  & -- & $\sim$2-3 min  \\
DreamBeast                                                   & 22.993          & 22.89          & 65          & 184.956          & 0.0935         & 0.76 & 18 & -- & $\sim$3-4 hr   \\
\midrule
Ours                                                         & \underline{28.170} & \textbf{19.26} & \textbf{84} & \underline{185.555} & \underline{0.051} & 0.86 & 18 & \textbf{798.31} & $\sim$3-5 min \\
\bottomrule
\end{tabular}
}
\label{tab:comparison}
\vspace{-3mm}
\end{table}

\vspace{-.5em}
\subsubsection{User Studies.}
\begin{figure}[t]
\vspace{-1mm}
  \centering
  \includegraphics[width=\columnwidth]{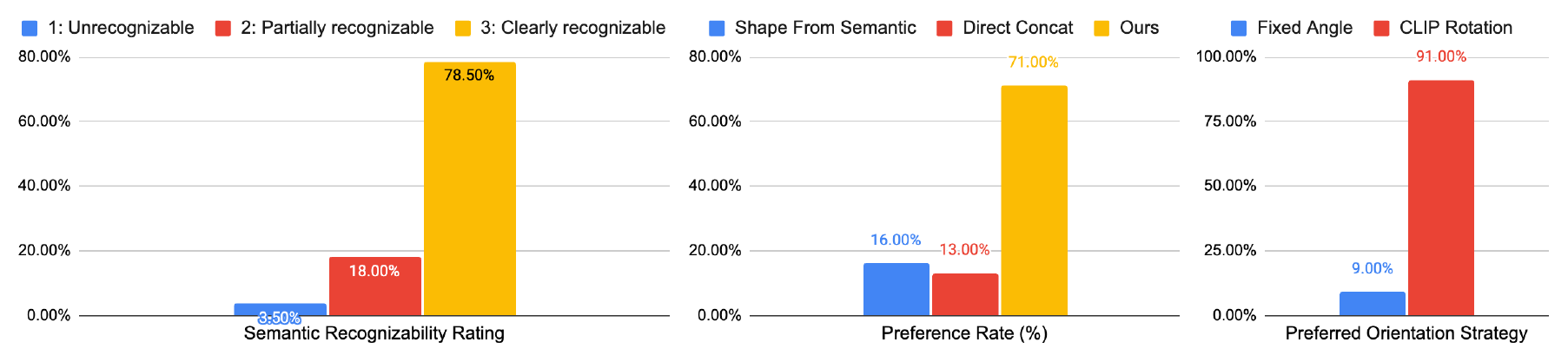}
\vspace{-6mm}
  \caption{
  \textbf{User study.}
  \textbf{(Left) Semantic Recognizability}: 78.5\% of participants rated our results as clearly recognizable (score 3).
  \textbf{(Middle) Method Preference}: 71\% of participants preferred our method over Shape From Semantics\cite{li2025shapesemantics3dshape} (16\%) and Direct Concatenation (13\%).
  \textbf{(Right) Orientation Strategy}: 91\% of participants found CLIP-guided orientation to produce a more natural illusion than fixed $0^\circ$/$180^\circ$ angles.
  }
  \vspace{-2mm}
  \label{fig:userstudy}
\vspace{-3mm}
\end{figure}

Our user studies strongly reinforce the quantitative findings. Participants overwhelmingly preferred our method over both baselines and rated our results as semantically recognizable, confirming the perceptual quality of our generated illusions. Additionally, the large majority found CLIP-guided orientation to produce more natural illusions than fixed angles, validating the effectiveness of our adaptive orientation search. Results are shown in Fig.~\ref{fig:userstudy}.

%左邊是語意辨識度，超過78%的參與者認為我們的結果是清晰可見的。中間則是超過70%的參與者在baseline中最喜歡我們的結果，右邊超過90%的參與者認為使用CLIP選擇的角度能夠得到更好的結果

%%%%%%%%%%%%%%%%%%%%%%%%%%%%%%%%%%%%%%%%%%%%%%%%%%
\vspace{-.5em}
\subsection{Ablation Studies}
%%%%%%%%%%%%%%%%%%%%%%%%%%%%%%%%%%%%%%%%%%%%%%%%%%

% \begin{figure*}[t]
%   \centering
%   \includegraphics[width=\textwidth]{figures/ablation_blend_method.pdf}
%   \caption{
%   \textbf{Ablation on geometry blending strategy} using the prompt pair \textit{``pumpkin carriage''} and \textit{``strawberry ice cream sundae in a cup''}.
%   Each row shows View~1, the blended mesh, and View~2.
%   (a)~\textbf{Union} produces conflicting geometry at the junction.
%   (b)~\textbf{Blur Average} yields smoother boundaries but loses geometric detail.
%   (c)~\textbf{Minkowski Blend} over-expands the geometry with limited detail preservation.
%   (d)~\textbf{Polar Coord Blend} introduces distortion for non-symmetric objects.
%   (e)~\textbf{SDF Average (Ours)} best balances geometric integrity and semantic preservation.
%   }
%   \label{fig:ablation_blend}
% \end{figure*}

\begin{figure}[t]
\vspace{-1mm}
  \centering
  \includegraphics[width=\columnwidth]{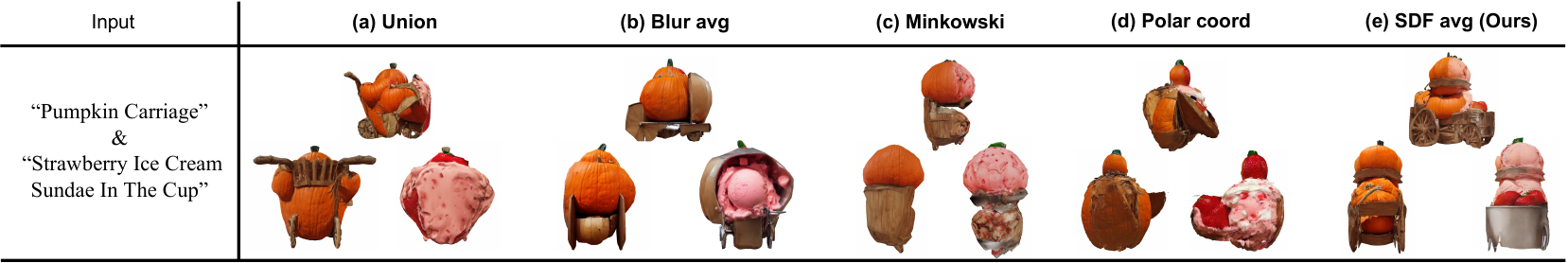}
  % \caption{
  % \textbf{Ablation on geometry blending strategy} using the prompt pair \textit{``pumpkin carriage''} and \textit{``strawberry ice cream sundae in a cup''}.
  % Each row shows View~1, the blended mesh, and View~2.
  % (a)~\textbf{Union} produces conflicting geometry at the junction.
  % (b)~\textbf{Blur Average} yields smoother boundaries but loses geometric detail.
  % (c)~\textbf{Minkowski Blend} over-expands the geometry with limited detail preservation.
  % (d)~\textbf{Polar Coord Blend} introduces distortion for non-symmetric objects.
  % (e)~\textbf{SDF Average (Ours)} best balances geometric integrity and semantic preservation.
  % }
  \vspace{-6mm}
  \caption{
  \textbf{Ablation on geometry blending} (\textit{``carriage''}/\textit{``sundae''}). Left-to-right: View 1, blended mesh, View 2. \textbf{(a) Union:} yields conflicting junctions. \textbf{(b) Blur Avg:} loses fine details. \textbf{(c) Minkowski:} over-expands geometry. \textbf{(d) Polar Coord:} distorts asymmetric objects. \textbf{(e) SDF Avg (Ours):} optimally balances geometric integrity and semantic preservation.
  % \textbf{Ablation on geometry blending strategy} (\textit{``pumpkin carriage''} \& \textit{``strawberry ice cream sundae in a cup''}).
  % Each row shows View~1, blended mesh, and View~2.
  % (a)~\textbf{Union}: conflicting geometry at the junction.
  % (b)~\textbf{Blur Average}: smoother boundaries but loses detail.
  % (c)~\textbf{Minkowski Blend}: over-expands the geometry.
  % (d)~\textbf{Polar Coord Blend}: distortion for non-symmetric objects.
  % (e)~\textbf{SDF Average (Ours)}: best balances geometric integrity and semantic preservation.
  }
  \label{fig:ablation_blend}
  \vspace{-3mm}
\end{figure}

\vspace{-.5em}
\subsubsection{Geometry Blending Strategy.}

We compare five voxel fusion strategies. \textbf{Union} takes the logical OR of two voxel grids, directly superimposing both structures without any shape coordination. \textbf{Blur Average} applies 3D Gaussian smoothing before element-wise averaging, introducing soft boundaries but losing fine geometric details. \textbf{Minkowski Blend} dilates each voxel with a spherical structuring element before merging, which tends to inflate the overall mesh volume. \textbf{Polar Coord Blend} fuses voxels slice-by-slice via 2D polar-coordinate boundary averaging, implicitly assuming star-shaped slices and thus failing for concave or non-symmetric objects. \textbf{SDF Average (Ours)} averages the Truncated SDFs of both occupancy grids and binarizes the result; the zero iso-surface naturally corresponds to the intermediate shape between the two objects, producing a stable and coherent blended surface. See Fig.~\ref{fig:ablation_blend} for visual comparisons.
\vspace{-4mm}

\begin{figure}[t]
\vspace{-1mm}
\centering

\begin{minipage}[t]{0.72\columnwidth}
\vspace{0pt}
    \centering
    \includegraphics[width=\linewidth]{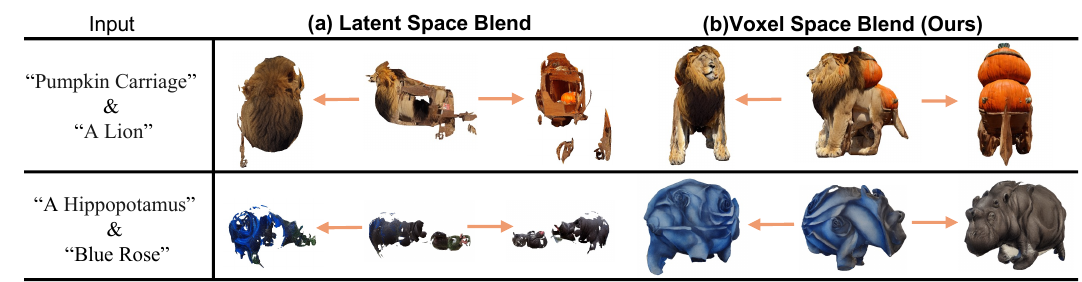}
\end{minipage}
\hfill
\begin{minipage}[t]{0.25\columnwidth}
\vspace{8pt}
    \centering
    \scriptsize
    \setlength{\tabcolsep}{2pt}
    \renewcommand{\arraystretch}{0.92}

    \resizebox{\linewidth}{!}{%
    \begin{tabular}{lcc}
    \toprule
    Method & Latent & Ours \\
    \midrule
    CLIP $\uparrow$      & 22.47 & \textbf{28.17} \\
    GPT $\uparrow$       & 58    & \textbf{84} \\
    FID $\downarrow$     & 195.6 & \textbf{185.6} \\
    KID $\downarrow$     & 0.081 & \textbf{0.051} \\
    Obj. Cnt $\downarrow$ & 1.30 & \textbf{0.86} \\
    Multi-Obj $\downarrow$ & 22  & \textbf{18} \\
    \bottomrule
    \end{tabular}}
\end{minipage}

\vspace{-2mm}
\caption{
\textbf{Ablation on geometry blending in latent vs.\ voxel space.}
\textbf{(a) Latent Space Blend}: averaging $z^1_{1|t}$ and $z^2_{1|t}$ directly in latent space without the decode--SDF blend--re-encode procedure falls outside TRELLIS's latent manifold, yielding incoherent and fragmented geometry.
\textbf{(b) Voxel Space Blend (Ours)}: SDF blending in voxel space produces geometrically coherent meshes with clear view-dependent semantics.
The quantitative comparison (right) confirms the necessity of our cross-space design.
}
% \vspace{-2mm}
\label{fig:ablation_latent_blend}
\vspace{-3mm}
\end{figure}

\vspace{-.5em}
\subsubsection{Latent Space vs.\ Voxel Space Blending.}
While the above strategies all operate in voxel space, one may ask whether
blending can be performed directly in latent space to avoid the
decode--re-encode overhead. We compare against \textbf{Direct Latent Averaging},
which averages $z^1_{1|t}$ and $z^2_{1|t}$ directly without the
decode--SDF blend--re-encode procedure. As shown in
Fig.~\ref{fig:ablation_latent_blend}, direct latent blending falls
outside TRELLIS's~\cite{xiang2025structured} latent manifold, producing
fragmented geometry and significantly worse quantitative performance
than our voxel-space blending, confirming the necessity of the proposed
cross-space design.
\vspace{-4mm}

\begin{figure*}[t]
\vspace{-1mm}
  \centering
  \includegraphics[width=\textwidth]{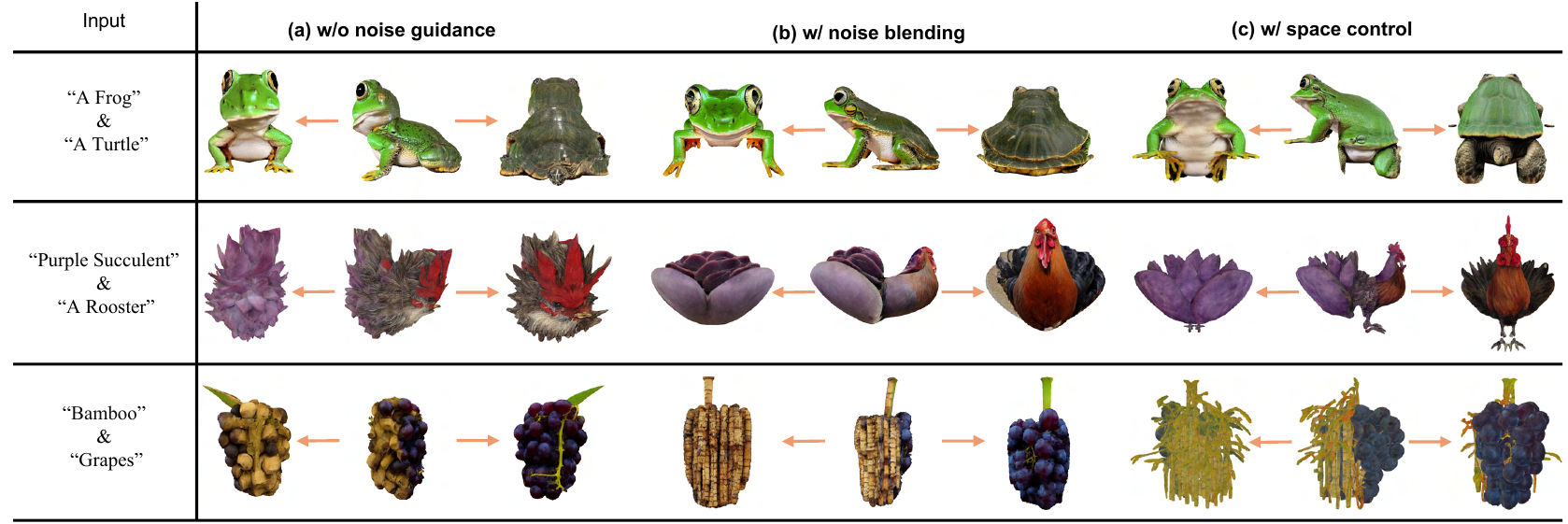}
\vspace{-6mm}
  \caption{
  \textbf{Ablation on noise guidance} across varying geometric compatibilities: (a)~no guidance, (b)~Noise Blending, and (c)~Space Control. \textbf{(c)} handles large geometric discrepancies (\textit{``Bamboo''/``Grapes''}) best via strong spatial constraints. \textbf{(b)} is optimal for similar silhouettes but distinct semantics (\textit{``Succulent''/``Rooster''}), avoiding semantic loss in (a) and residual artifacts in (c). For compatible pairs (\textit{``Frog''/``Turtle''}), all settings perform comparably. Thus, optimal intervention strength depends on geometric alignment.
  % \textbf{Ablation on noise guidance strategy} across three prompt pairs with varying geometric compatibility.
  % Each column shows results under (a)~no guidance, (b)~Noise Blending Guidance, and (c)~Space Control Guidance.
  % Space Control best handles pairs with large geometric discrepancies (\textit{``Bamboo''} \& \textit{``Grapes''}) by strongly constraining the initial spatial distribution.
  % Noise Blending performs best for pairs with similar silhouettes but distinct semantics (\textit{``Purple Succulent''} \& \textit{``A Rooster''}): w/o guidance fails to surface the rooster, while Space Control over-constrains and leaves residual rooster geometry in the succulent viewpoint.
  % For compatible pairs (\textit{``A Frog''} \& \textit{``A Turtle''}), all settings produce comparable results.
  % Higher intervention does not always lead to better results.
  }
  \label{fig:ablation_noise}
\vspace{-3mm}
\end{figure*}

\vspace{-.5em}
\subsubsection{Noise Guidance.}
We compare three configurations: no guidance, Noise Blending Guidance, and Space Control Guidance. The optimal strategy depends on the geometric characteristics of the object pair: Space Control imposes stronger constraints via the guided latent state for the first $t_0$ steps, making it better suited for pairs with large silhouette discrepancies; Noise Blending provides a milder prior that benefits pairs with similar silhouettes but distinct semantics; for geometrically compatible pairs, no guidance suffices. We therefore treat noise guidance as an optional, pair-dependent auxiliary rather than a mandatory component. Qualitative comparisons are shown in Fig.~\ref{fig:ablation_noise}.

\begin{figure*}[t]
\vspace{-1mm}
  \centering
  \includegraphics[width=\textwidth]{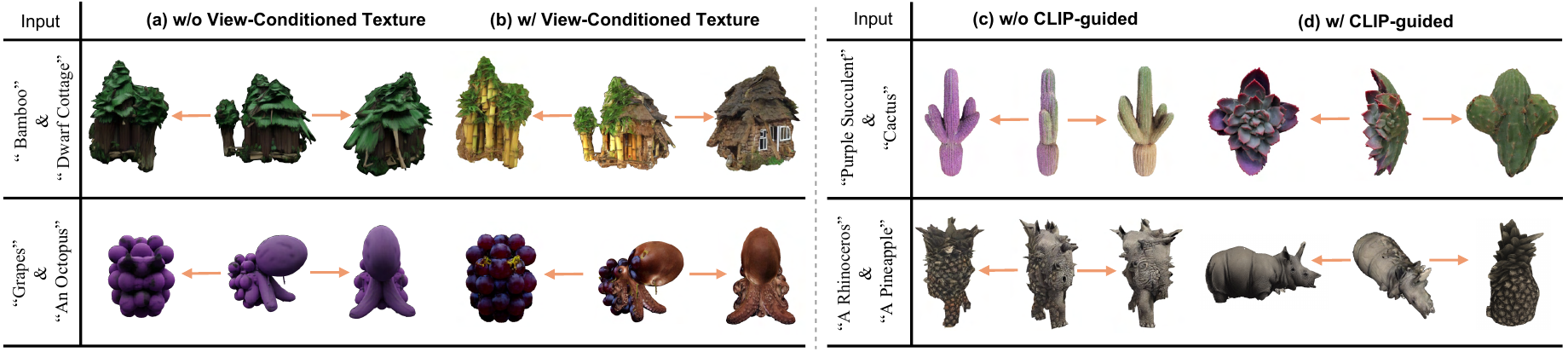}
  \vspace{-6mm}
  \caption{
  \textbf{Ablation on texture synthesis and orientation search.}
  Each pair shows View 1, blended mesh, and View 2. \textbf{(a) w/o View-Conditioned Texture:} Standard texturing fails on fused geometries, blending semantics into indistinguishable appearances (e.g., uniform foliage or solid purple). \textbf{(b) w/ View-Conditioned Texture (Ours):} Accurately assigns distinct, prompt-driven textures to each viewpoint. \textbf{(c) w/o CLIP-guided:} Fixed $0^\circ$/$180^\circ$ angles cause silhouette misalignments, failing to reveal clear semantics. \textbf{(d) w/ CLIP-guided (Ours):} Adaptive rotation optimally aligns silhouettes, recovering distinct semantics at both views.
  \vspace{-2mm}
  % \textbf{Ablation on view-conditioned texture synthesis and CLIP-guided Orientation Search.}
  % Each pair shows View~1, blended mesh, and View~2.
  % \textbf{(a)~w/o View-Conditioned Texture}: texturing via TRELLIS fails on the unnatural fused geometry, producing uniform foliage for \textit{``Bamboo''} \& \textit{``Dwarf Cottage''} and a single purple tone for \textit{``Grapes''} \& \textit{``An Octopus''}, rendering both semantics indistinguishable.
  % \textbf{(b)~w/ View-Conditioned Texture (Ours)}: view-conditioned synthesis assigns each viewpoint its own semantically driven texture, closely matching the corresponding prompt.
  % \textbf{(c)~w/o CLIP-guided}: fixed $0^\circ$/$180^\circ$ angles cause silhouette misalignment for both pairs, failing to present distinct semantics at each viewpoint.
  % \textbf{(d)~w/ CLIP-guided (Ours)}: adaptive rotation recovers distinct semantics at both viewpoints for both pairs.
  }
  \label{fig:ablation_texture_clip}
\end{figure*}

\vspace{-.5em}
\vspace{-3mm}
\subsubsection{View-Conditioned Texture Synthesis.}
We ablate the necessity of Stage~2 by comparing results with and without view-conditioned texturing (Fig.~\ref{fig:ablation_texture_clip}(a)(b)). Without Stage~2, TRELLIS fails to interpret the unnatural fused geometry, producing semantically incoherent textures at both viewpoints. Our view-conditioned synthesis assigns each viewpoint its own texture prediction, ensuring semantic consistency with the target prompt.

\vspace{-.5em}
\vspace{-3mm}

\subsubsection{CLIP-Guided Orientation Search.}

We compare adaptive CLIP-guided rotation against fixed $0^\circ$/$180^\circ$ angles (Fig.~\ref{fig:ablation_texture_clip}(c),(d)). For canonically misaligned pairs (e.g., horizontal \textit{``Rhinoceros''}/upright \textit{``Pineapple''}), fixed angles cause silhouette misalignment; CLIP rotation optimally aligns them to recover distinct semantics. It similarly prevents semantic collapse for \textit{``Succulent''/``Cactus''}. However, for already compatible pairs, CLIP misclassifications can yield suboptimal angles. Thus, we recommend adaptive rotation specifically for geometrically challenging pairs, defaulting to fixed angles for compatible ones.
% \vspace{-2mm}
% We compare adaptive CLIP-guided angle selection against fixed $0^\circ$/$180^\circ$ angles (Fig.~\ref{fig:ablation_texture_clip}(c)(d)). For pairs with substantially different canonical orientations---e.g., \textit{``A Rhinoceros''} (horizontal) paired with \textit{``A Pineapple''} (upright)---fixed angles cause silhouette misalignment; CLIP Rotation automatically selects the best-aligned orientation, recovering distinct semantics at both viewpoints. Similarly, fixed angles collapse \textit{``Purple Succulent''} \& \textit{``Cactus''} to a cactus-like appearance, while CLIP Rotation recovers the succulent semantics. However, for geometrically compatible pairs, CLIP's non-negligible misclassification rate may select a suboptimal angle, making fixed angles more stable. We therefore recommend CLIP Rotation for geometrically challenging pairs and fixed angles for compatible ones.

\vspace{-.5em}
\subsection{Applications}
\label{sec:applications}
\vspace{-.5em}

\begin{figure}[t]
\vspace{-1mm}
  \centering
  \includegraphics[width=\columnwidth, height=0.3\textheight, keepaspectratio]{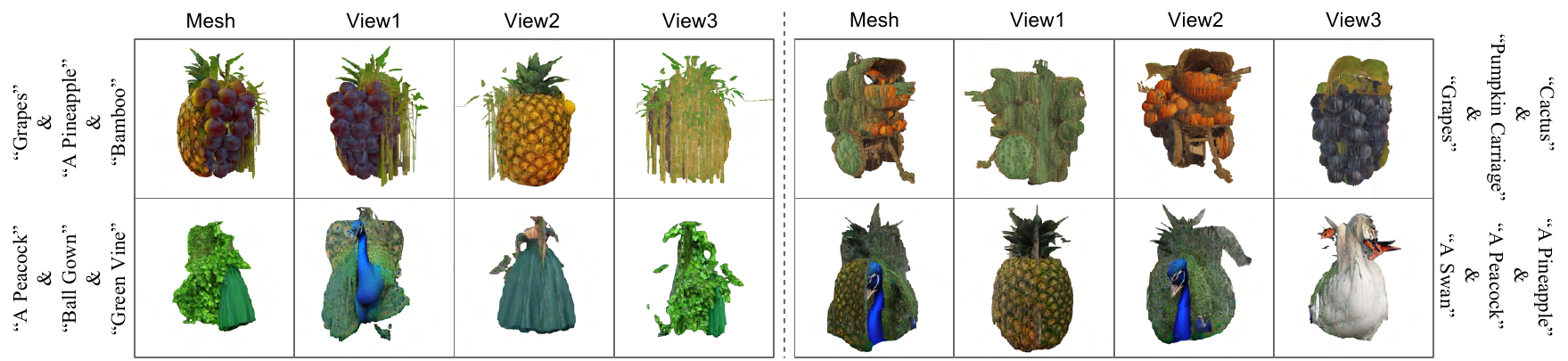}
  \vspace{-6mm}
  \caption{
  \textbf{Qualitative results of three-object 3D illusion generation.}
  Each row shows the fused mesh and three target-viewpoint renders for a prompt triplet.
  Our method successfully generates a single coherent mesh that presents three distinct semantics at $0^\circ$, $120^\circ$, and $240^\circ$, respectively.
      }
    \vspace{-1mm}
  \label{fig:qualitative_three_object}
\vspace{-3mm}
\end{figure}

Our framework demonstrates strong performance across diverse semantic combinations, from structurally similar to geometrically distinct object pairs, as shown in Fig.~\ref{fig:qualitative_fixed} and Fig.~\ref{fig:qualitative_clip}. Beyond the two-object case, our method scales naturally to three-object illusion generation without architectural modifications (Sec.~\ref{sec:three_object}), as demonstrated in Fig.~\ref{fig:qualitative_three_object}.

% \noindent\textbf{Three-Object 3D Illusion.}
% Our framework naturally scales to three-object illusion generation without architectural modifications, as described in Sec.~\ref{sec:three_object}. Qualitative results on three prompt triplets are shown in Fig.~\ref{fig:qualitative_three_object}, demonstrating that the generated mesh presents three distinct and recognizable semantics at the three target viewpoints.

\vspace{-.5em}
\section{Conclusion}
\vspace{-.5em}
We present a zero-shot framework that extends visual illusions to true 3D geometry. Given text prompts, our method generates a single coherent mesh revealing distinct semantics from different viewpoints in under 5 minutes, without per-shape optimization. Furthermore, we introduce CLIP-guided Orientation Search for silhouette alignment and Noise Guidance to resolve geometric conflicts, demonstrating seamless scalability to three-object illusions without modifying the core fusion procedure.
% We present a zero-shot framework for generating 3D visual illusions, where a single 3D mesh reveals semantically distinct appearances at different target viewpoints, extending visual illusions from 2D images to true 3D geometry. Given two text prompts, our method produces a coherent 3D object that embodies dual semantics without any per-shape optimization, completing within 5 minutes. We further propose CLIP-guided Orientation Search for automatic silhouette alignment, two Noise Guidance strategies to resolve geometric conflicts between semantics, and demonstrate scalability to three-object illusion generation---all without modifying the core fusion procedure.

\vspace{-.5em}
\subsubsection{Limitations.}

Our method has three limitations. First, it inherits failure cases from TRELLIS~\cite{xiang2025structured}; for example, TRELLIS may misinterpret \textit{``bat''} as Batman, yielding incompatible geometries. Second, our framework requires a voxel-based latent representation and is therefore not directly applicable to token-based 3D generators such as Hunyuan3D~\cite{zhao2025hunyuan3d}. Third, CLIP-guided Orientation Search scales poorly to three-object illusions due to ambiguous silhouette alignment, so we use fixed target angles ($0^\circ$, $120^\circ$, $240^\circ$).

\bibliographystyle{splncs04}
\bibliography{main}
\end{document}